\documentclass[a4paper]{article}

\usepackage[hyperref, acceptedwithA]{tacl2018v2}
\usepackage{bm}
\usepackage{times, latexsym}

\usepackage{graphicx}
\usepackage{array}
\newcolumntype{?}{!{\vrule width 1pt}}
\usepackage[symbol]{footmisc}
\usepackage{booktabs}       
\usepackage[utf8]{inputenc} 
\usepackage[T1]{fontenc}    
\usepackage{hyperref}       
\usepackage{url}            
\usepackage{amsfonts}       
\usepackage{nicefrac}       
\usepackage{microtype}      
\usepackage{amsmath}
\usepackage{amssymb}

\usepackage{bbm}
\usepackage{comment}
\usepackage{multicol}
\usepackage{appendix}
\usepackage{float}
\usepackage{threeparttable}
\usepackage{lmodern,textcomp}
\usepackage{breqn}
\usepackage{subcaption}
\usepackage{algorithm}
\usepackage[noend]{algpseudocode}
\algnewcommand{\LineComment}[1]{\State \(\triangleright\) #1}

\makeatletter
\def\BState{\State\hskip-\ALG@thistlm}
\makeatother

\newcommand{\norm}[1]{\left\lVert#1\right\rVert}

\title{Efficient Content-Based Sparse Attention with Routing Transformers}
\author{Aurko Roy \and Mohammad Saffar \and Ashish Vaswani \and David Grangier \\
Google Research\\ \texttt{\{aurkor, msaffar, avaswani, grangier\}@google.com} }
\begin{document}
\maketitle
\begin{abstract}
Self-attention has recently been adopted for a wide range of sequence modeling problems.
Despite its effectiveness, self-attention suffers from quadratic compute and memory requirements with respect to sequence length. Successful approaches to reduce this complexity focused on attending to local sliding windows or a small set of locations \emph{independent} of content. 
Our work proposes to learn dynamic sparse attention patterns that avoid allocating 
computation and memory to attend to content unrelated to the query of interest.
This work builds upon two lines of research: it combines the modeling flexibility of 
prior work on \emph{content-based} sparse attention with the efficiency gains from 
approaches based on \emph{local, temporal} sparse attention. Our model, the Routing 
Transformer, endows self-attention with a sparse routing module based on online \(k\)-means
while reducing the overall complexity of attention to \(O(n^{1.5}d)\) from \(O(n^2d)\)
for sequence length \(n\) and hidden dimension \(d\).
We show that our model outperforms comparable sparse attention models
on language modeling on \texttt{Wikitext-103} ($15.8$ vs $18.3$ perplexity),
as well as on image generation on \texttt{ImageNet-64} ($3.43$ vs $3.44$ bits/dim)
while using fewer self-attention layers. Additionally, we set a new state-of-the-art 
on the newly released \texttt{PG-19} data-set, obtaining a test perplexity of 
\(33.2\) with a \(22\) layer Routing Transformer model trained on sequences of length \(8192\). We open-source the code for Routing Transformer in Tensorflow.
\footnote[1]{\url{https://github.com/google-research/google-research/tree/master/routing_transformer}}
\end{abstract}

\section{Introduction}

Generative models of sequences have witnessed rapid progress
driven by the application of attention to neural networks. In 
particular, \citet{bahdanau2014neural,cho2014learning,vaswani2017attention}
relied on attention to drastically improve the state-of-the art
in machine translation. Subsequent research
\citep{radford2018improving,devlin2018bert,liu2019multi, yang2019xlnet}
demonstrated the power of self-attention in learning powerful 
representations of language to address several natural language
processing tasks.
Self-attention also brought impressive progress for generative modeling 
outside of language, e.g. image \citep{parmar2018image, menick2018generating, child2019generating} and
music generation~\citep{huang2018music, child2019generating}.

Self-attention operates over sequences in a step-wise manner: at 
every time-step, attention assigns an \emph{attention weight} to each 
previous input element (representation 
of past time-steps) and uses these weights to compute the representation of the current 
time-step as a weighted sum of the past input elements~\citep{vaswani2017attention}. 
Self-attention~\citep{shaw2018self} is a particular case of attention~\citep{bahdanau2014neural,chorowski2015attention,luong2015effective}.

Self-attention is commonly used in auto-regressive generative models.
These models generate observations step-by-step, modeling the probability 
of the next symbol given the previously generated ones. At every time step, 
self-attentive generative models can directly focus on any part of the 
previous context. In contrast, recurrent neural networks (RNNs) and 
convolutional neural networks (CNNs) have direct interactions with only a 
local neighborhood of context around the current time step. 

This advantage however comes at a price: unlike recurrent networks or 
convolution networks, the time and space complexity of self-attention 
is quadratic in $n$, the length of the sequence. Specifically, for every
position $i \le n$, self-attention computes weights for its whole context of 
length $i$, which induces a complexity of $\sum_{i \le n} i = n (n-1)/2$.
This makes it difficult to scale attention based models 
to modeling long sequences. However, long sequences 
are the norm in many domains, including music, image, speech, video generation
and document level machine translation.

Therefore, an important research direction is to investigate sparse and 
memory efficient forms of attention in order to scale to tasks with large 
sequence lengths.
Previous work has proposed \emph{data independent} or fixed sparsity patterns bounding 
temporal dependencies, such as local or strided attention. At each time step, the model 
attends only to a fixed number of time steps in the past \citep{child2019generating}. 
Extensions to local attention have suggested learning the length of the temporal sparsity 
for each attention module in the network~\citep{sukhbaatar2019adaptive}. 
These strategies draw their inspiration from RNNs and CNNs and bound their 
complexity by attending only to representations summarizing a \emph{local}
neighborhood of the current time step. Their attention matrices (matrices containing
the attention weights for every pair of previous, current time-step) are natively sparse
and require instantiating only non-zero entries. While these approaches have achieved good results, fixing the sparsity pattern of a content based mechanism such as self-attention can limit its ability to pool in information from large contexts. 

As an alternative to local attention, \citet{correia2019adaptively} 
consider content-based sparsity, an approach allowing for arbitrary 
sparsity patterns. This formulation however does require instantiating a full 
dense attention matrix prior to sparsification through variants of \(L_0\)-sparsity 
or sparsemax approximations~\citep{blondel2019Fenchel}.

The present work builds upon these two lines of research and proposes to retain 
the modeling flexibility of content-based sparse attention while
leveraging the efficiency of natively sparse attention matrices.
Our formulation avoids sparsemax variants and relies on clustering of attention 
instead. Each attention module considers a clustering of the space: the
current time-step only attends to context belonging to the same cluster.
In other words, the current time-step query is \emph{routed} to a limited
number of context elements through its cluster assignment. This strategy draws inspiration
from the application of spherical \(k\)-means clustering to the Maximum Inner
Product Search (MIPS) problem.

Our proposed model, Routing Transformer, combines our efficient clustering-based 
sparse attention with classical local attention to reach excellent performance 
both for language and image generation. These results are obtained without the 
need to maintain attention matrices larger than batch length which is the case 
with the segment level recurrence mechanism used in~\citet{dai2019transformer,
sukhbaatar2019adaptive}. 
We present experimental results on language modeling 
(\texttt{enwik-8}, \texttt{Wikitext-103} and \texttt{PG-19}) and unconditional image generation (\texttt{CIFAR-10} and \texttt{ImageNet-64}). 
Routing Transformer sets new state-of-the-art while having comparable or fewer 
number of self-attention layers and heads, on 
\texttt{Wikitext-103} ($15.8$ vs $18.3$ perplexity),
\texttt{PG-19} (\(33.2\) vs \(33.6\) perplexity),
and on \texttt{ImageNet-64} ($3.43$ vs $3.44$ bits/dim). We also report
competitive results on \texttt{enwik-8} ($0.99$ vs $0.98$ perplexity)
and present ablations on \texttt{CIFAR-10}.

\section{Related Work}

{\bf Attention with Temporal Sparsity:} 
Research on efficient attention neural models parallels the advent of
attention-based architectures.
In the context of speech recognition, \citet{jaitly2016online} proposed the 
Neural Transducer which segments sequences in non-overlapping chunks and 
attention is performed in each chunk independently. Limiting attention to a 
fixed temporal context around the current prediction has also been explored in
\citet{chorowski2015attention}, while \citet{chiu2018monotonic} dynamically 
segment the sequence into variable sized-chunks.

Hierarchical attention strategies have also been explored: the model first considers
which part of the inputs should be attended to before computing full attention in a
contiguous neighborhood of the selected area~\citep{gregor2015draw,vinyals2015show,luong2015effective}.
Later, hierarchical attention has been simplified by \citet{liu2018generating} that 
alternates coarse layers (attending to the whole sequence at a lower temporal resolution) 
with local layers (attending to a neighborhood of the current prediction).

This alternating strategy is also employed by \citet{child2019generating}, which
introduces bounded and strided attention, i.e. attending to a fixed context in the past
at a sub-sampled temporal resolution. This work formalizes such a strategy using a
sparse attention formalism, showing how it relates to full attention with
a specific sparsity pattern in the attention matrix. It shows that sparse attention 
is sufficient to get state-of-the-art results in modeling long sequences over
language modeling, image generation and music generation. \citet{sukhbaatar2019adaptive}
build upon this work and show that is it is possible to obtain further sparsity
by letting the model learn the length of the temporal context for each attention module. 
This work also makes use of the attention cache introduced in 
\citet{dai2019transformer}, a memory mechanism to train models over 
temporal contexts which extend beyond the length of the training batches.

{\bf Attention with Content-Based Sparsity:} 
The above work mainly relies on two efficient ideas: attending to 
less elements by only considering a fixed bounded local context in the past, and 
attending to less elements by decreasing the temporal resolution of context.
These ideas do not allow arbitrary sparsity patterns in attention matrices. Content-based
sparse attention has been introduced to allow for richer patterns and more expressive models.
\citet{martins-kreutzer-2017-learning, malaviya-etal-2018-sparse} propose to compute attention
weights with variants of sparsemax. \citet{correia2019adaptively} generalizes this approach to
every layer in a Transformer using entmax which allows for more efficient inference.
This line of work allows for learning arbitrary sparsity attention patterns from data, based on
the content of the current query and past context. However, sparsity here cannot be leveraged
to improve space and time complexity since sparsemax/entmax formulations require instantiating
the full attention matrix prior to sparsification. This is a drawback compared to temporal
sparsity approaches. Our work is motivated by bridging this gap and allows for arbitrary sparsity
patterns while avoiding having to instantiate non-zero entries of attention matrices.

Contemporaneous to our work, \citet{kitaev2020reformer} proposed to use Locality Sensitive Hashing (LSH)
using random hyper-planes to infer content based sparsity patterns for attention: tokens that fall into the same
hash bucket, get to attend to each other. While similar in spirit
to our approach, the approach of \citet{kitaev2020reformer} keeps the randomly initialized hyper-planes
fixed throughout, while we use mini-batch spherical \(k\)-means to learn the space-partitioning centroids.
The motivation in both approaches is to approximate Maximum Inner Product Search (MIPS) in the context of
dot product attention, for which both LSH and spherical \(k\)-means have been used in literature. However, 
typically spherical \(k\)-means is known to out-perform LSH for MIPS (see e.g. \citet{auvolat2015clustering}).
This is borne out in the common task of \texttt{Imagenet-64} generation, 
where Reformer gets around \(3.65\) bits/dim (Figure 3), while the Routing Transformer gets \(3.43\) bits/dim (see Table~\ref{tab:imagenet} for a comparison).

{\bf Sparse Computation beyond Attention:}
Learning models with sparse representations/activations for saving time and computation
has been addressed in the past in various context. Previous work often refers to this goal as
\emph{gating} for conditional computation. Gating techniques relying on sampling and 
straight-through gradient estimators are common~\citep{bengio2013estimating,eigen2013learning,cho2014exponentially}. 
Conditional computation can also be addressed with reinforcement learning 
\citep{denoyer2014deep, indurthi2019look}. Memory augmented neural networks 
with sparse reads and writes have also been proposed in \citet{rae2016scaling}
as a way to scale Neural Turing Machines~\citep{graves2014neural}.
In the domain of language modeling, 
a related work is the sparsely gated Mixture-of-experts (MOE)  
\citep{shazeer2017outrageously} where sparsity is induced by \emph{experts} 
and a trainable gating network controls the routing strategy to each 
sub-network. Another related work is \citet{lample2019large} who use product
quantization based key-value lookups to replace the feed forward network
in the Transformer. Our work differs from theirs in that we make use of dynamic
key-value pairs to infer sparsity patterns, while their key-value pairs are
the same across examples.
\section{Self-Attentive Auto-regressive Sequence Modeling}
\label{sec:attention}

Auto-regressive sequence models decompose the probability of 
a sequence \(\mathbf{x} = \left(x_1, \dots, x_n\right)\) as
\begin{align}
    p(\mathbf{x}) = p_\theta(x_1) \prod_{i=2}^{n} p_\theta(x_{i} | x_{< i}).
\end{align}
In neural models, the conditional distribution 
\(p_\theta(x_{i} | x_{< i})\) 
is modeled by a neural network with learned parameters \(\theta\)
and these parameters are typically learned to maximize the 
likelihood of the training data.
In particular, Transformer architectures have shown to reach 
state-of-the-art accuracy in several domains,
including language modeling \citep{vaswani2017attention, radford2018improving}, 
image generation \citep{parmar2018image} and 
music generation \citep{huang2018music}. 
Transformer models compose a series of attention modules. Each module
refines the input representation by taking a weighted average of the 
representations from the previous modules.

For every module, the input representation is a sequence of $n$ vectors
$
\mathbf{x} = (x_1, \dots, x_n)
$
from a continuous space of dimension \(d\). Thus one may actually treat 
the input sequence as a \(n \times d\) matrix \(X\). A self-attention layer
operates on this representation. It first applies three linear projections,
\begin{align}
Q = X W_Q, \quad
K = X W_K, \quad
V = X W_V,
\end{align}
where $Q, K$ and $V$ are referred to as \emph{keys}, \emph{queries} and \emph{values},
while $W_Q, W_K, W_V$ are learned projection matrices.

The key and the query matrices determine the \(n \times n\) 
attention matrix \(A = \operatorname{softmax}\left(QK^\top\right)\),
where the softmax operator over matrices denotes that the softmax function 
has been applied to each row. 
In the case of self-attention for auto-regressive models, queries attend
only over keys from previous time-steps, i.e. 
\begin{align}
A = \operatorname{softmax}\left(\operatorname{ltr}(QK^\top\right))
\end{align}
where $\operatorname{ltr}$ denotes the lower triangular operator.
The attention matrix \(A\) may be interpreted as a matrix of weights 
in \([0, 1]\) where \(A_{ij}\) denotes how much query position \(i\) at the next 
layer must pay attention to key position \(j\) at the previous layer. 
Given the attention matrix \(A\), the next layer representation \(X'\) 
is then computed simply as \(A V\). In summary, 
\begin{align}
X'_i =  \sum_{j < i}^n A_{ij}V_j,
\end{align}
In practice, Transformer \citep{vaswani2017attention} adds several extensions to
this basic self-attention mechanism. 
In particular, the result \(X'\) of performing self-attention is scaled by
\(1/\sqrt{d}\).
Moreover, each layer relies on 
multiple attention \emph{heads}, i.e. each layer performs multiple projections 
onto triplet (queries, keys, values) and attention is performed for each head. 
The attention results from all heads are then concatenated. This strategy allows each 
head 
to specialize on different aspects of the input sequence. In addition, Transformer
further processes the result of attention through a learnable non-linear 
transformation 
(multi-layer perceptron, $\operatorname{mlp}$) followed by a residual connection and 
a normalization step, i.e.
\begin{align}
X' &= \operatorname{layernorm}(X' + X)\\
X'' &= \operatorname{layernorm}(\operatorname{mlp}(X') + X),
\label{eq:attn_block}
\end{align}
where $\operatorname{layernorm}$ denotes the parameterized normalization step 
from \cite{layernorm2016}. A full Transformer model is therefore a chain of
attention modules (Eq.~\ref{eq:attn_block}) preceded by an embedding 
module (learnable representation for symbols and their positions) and followed
by a logistic classification module (learnable linear classifier to predict
the next symbol).

Our work is interested in the application of the Transformer to long sequences,
a challenging problem since space and time complexity of attention is quadratic 
in sequence length \(n\). We describe various approaches to sparse attention 
including ours in the next section.

\section{Efficient Content-Dependent Sparse Attention}\label{sec:ksparse}
Attention-based models can be problematic for long sequences. For a sequence of 
length $n$, the full attention matrix $A$, as introduced in Section~\ref{sec:attention}, 
is \(n \times n\)-dimensional and can be prohibitive to instantiate. 
This motivates sparse attention models, i.e. models relying on attention matrices
which have a majority of zero entries. 

For each query, a sparse attention model defines a set of keys which can be attended to.
In the following, we introduce the set $S_i$ as the set of key positions that the query at 
position $i$ can attend to, i.e.
\begin{align}
X'_i = 
    \sum_{j \in S_i} A_{ij} V_j.
\end{align}
The set of all such key positions defines a sparsity pattern \(\mathcal{S} = \{S_i \mid 
1 \le i \le n\}\) for the entire sequence.
For example, classical causal self attention can attend to every key prior to the current
query, which translates to $S_i = \{ j \mid j < i\}$ for every \(i\).
Most previous work on attention sparsity defined such sets purely based on positions, 
independently of actual query and key vectors. For example, local attention \citep{luong2015effective} considers attending only to a $k$-long time window 
prior to the current query, 
\(S_i = \{j \mid i - k \le j < i\}\) for every \(i\). 
The work of \citet{child2019generating} propose block
sparse attention where half the heads perform local attention, and half the heads perform \emph{strided attention} given by \(S_i = \{j \mid i - j \pmod k = 0, j < i\}\)
for every \(i\). 
The approach of \citet{sukhbaatar2019adaptive} is also a variant of local attention where the 
cardinality of $|S_i|$ is learned from data with an $L_1$ penalty to trade-off 
sparsity with modeling accuracy.

These \emph{local} attention sparsity variants are effective in practice since
correlation between observations naturally decrease with time for many problems.
In our experiments, we actually find that local attention is a surprisingly 
strong baseline in both image generation and language modeling: for e.g., a 
scaled up ImageTransformer \citep{parmar2018image}  gets \(3.48\) bits/dim 
compared to the \(3.44\) bits/dim reported in \citet{child2019generating}.
Similarly, scaled up versions of Transformer with local attention and the relative
positional encoding scheme of \citet{shaw2018self} are able to get \(19.8\) perplexity on 
\texttt{Wikitext-103}, \(1.10\) bits per byte on \texttt{enwik-8} and \(39.3\) on \texttt{PG-19},
while Transformer-XL \citep{dai2019transformer} gets \(18.3\), \(0.99\) and \(36.3\) respectively. 
From an efficiency perspective, local attention is
also interesting since sparsity patterns are regular, contiguous in memory and 
known in advance.

In this work, however, we are interested in a more generic formulation of attention
sparsity and would like the sparsity pattern to be informed by the data, 
i.e., \(\mathcal{S} = f(\mathbf{x})\). 
This approach has several modeling advantages: it can accommodate data without 
a clear ordering over observations. For temporal data, it can also discover 
patterns with greater sparsity if some types of queries have a longer lasting 
effect on future observations than others.
Content-based sparse attention should however be carefully implemented if we
need to avoid instantiating full attention matrices at any point in time.
For instance, \citet{correia2019adaptively} infer sparsity from data but their 
formulation instantiates a full attention matrix before finding its sparse counterpart.
The next section explains how a natively sparse approach can actually be devised
inspired by the Maximum Inner Product Search (MIPS) problem.

\subsection{Routing Attention with Clustering}\label{sec:routing}
Our strategy follows the motivation we delineated in the previous section:
we model sparse attention matrices with a low rank sparsity patterns relying 
on \(k\)-means clustering. Our strategy first assigns queries and keys to clusters.
Then only queries and keys from the same cluster are considered for attention.

Precisely, our model clusters both keys \(K\) and queries \(Q\) 
using mini-batch \(k\)-means clustering on the same
set of centroid vectors 
$\bm{\mu} = (\mu_1, \cdots, \mu_k) \in \mathbb{R}^{k \times d}$. These
centroid parameters are model parameters and are shared across 
sequences. They are learned online along with the rest of the parameters, as 
delineated in \cite{bottou1995convergence}. 
Once cluster membership for queries and keys are determined, we
denote by \(\mu(Q_i)\in \bm{\mu}\) the nearest centroid to \(Q_i\) and
by \(\mu(K_j) \in \bm{\mu}\) the nearest centroid to \(K_j\).
This allows us to define our sparse attention strategy as
\begin{align}\label{eq:high-res}
    X'_i = \sum_{\substack{j: K_j \in \mu(Q_i),\\ j <i}} A_{ij}V_j
\end{align}
In summary, queries are routed to keys belonging to the same cluster.
To see the connection with Maximum Inner Product Search (MIPS), we recall
the setting of the MIPS problem adapted to the case of dot-product attention. 
In this problem we are given a large collection of vectors 
\(\mathcal{K} = \{K_1,\cdots, K_n\}\) of size \(n\) in \(\mathbb{R}^d\)
and for a given query \(Q_i \in \mathbb{R}^d\),
we are interested in searching for a key \(K_j \in \mathcal{K}\) which (approximately)
maximizes \(Q_i^\top K_j\):
\begin{align}\label{eq:mips}
    K_j = \arg\max_{x \in \mathcal{K}} Q_i^\top x.
\end{align}
The MIPS problem is useful in the dot product attention setting because the importance
of a particular key \(K_j\) to a query \(Q_i\) is directly proportional to its
dot product \(Q_i^\top K_j\). Thus given a budget of items that a query
\(Q_i\) can attend to, the optimal choice of keys \(K_j\) are the ones given by
the MIPS objective in Equation~\ref{eq:mips}. 
The motivation for using \(k\)-means clustering, is the
observation that the MIPS problem is equivalent to the Nearest Neighbor
Search (NNS) problem when the norm of every element \(K_j \in \mathcal{K}\) is
constant. 

Therefore, we work with queries and keys 
which are unit vectors, projecting them onto the
unit ball, immediately before computing them.
In practice, instead of normalizing by the \(\ell_2\) norm, we 
use Layer Normalization \citep{layernorm2016} with the scale and bias
terms disabled. This has the benefit of projecting vectors in \(\mathbb{R}^d\) to the 
\(d\)-ball and prevents its entries from becoming too small. 
These layer normalized keys and queries are also
used subsequently for computing the dot product attention.
Note that performing \(k\)-means algorithm on unit vectors 
is equivalent to the \emph{spherical} \(k\)-means algorithm. 
Projecting queries and keys to the unit ball implies that:
\begin{align}
    &\norm{Q_i - K_j}^2\\
    &= \norm{Q_i}^2 + \norm{K_j}^2 - 2Q_i^\top K_j \\
    &=  2 - 2\left(Q_i^\top K_j\right).\label{eq:dist-preserving}
\end{align}
Thus if \(Q_i\) and \(K_j\) belong to the same cluster center 
i.e., \(\mu(Q_i) = \mu(K_j) = \mu\),
then it follows that there is some \(\varepsilon > 0\), such
that \(\norm{Q_i - \mu}, \norm{K_j - \mu} < \varepsilon\).
This implies via triangle inequality that:
\begin{align}
    \norm{Q_i - K_j} \le \norm{Q_i - \mu} + \norm{K_j - \mu} < 2\varepsilon.
\end{align}
Thus from Equation~\ref{eq:dist-preserving} it follows that,
\(Q_i^\top K_j > 1 - 2\varepsilon^2\).
Therefore, when two time steps $i > j$ are assigned the same cluster due to 
a small $\norm{Q_i - \mu}, \norm{K_j - \mu}$ distance, 
it also means that their attention weight
$Q_i^\top K_j$ is high, i.e., \(K_j\) is an approximate solution
to the MIPS objective of Equation~\ref{eq:mips} for query \(Q_i\). 
This analysis shows that our clustering routing strategy 
preserves large attention weights as non-zero entries.

\begin{figure*}[h]
\begin{subfigure}{.33\textwidth}
  \centering
  \includegraphics[width=.8\linewidth]{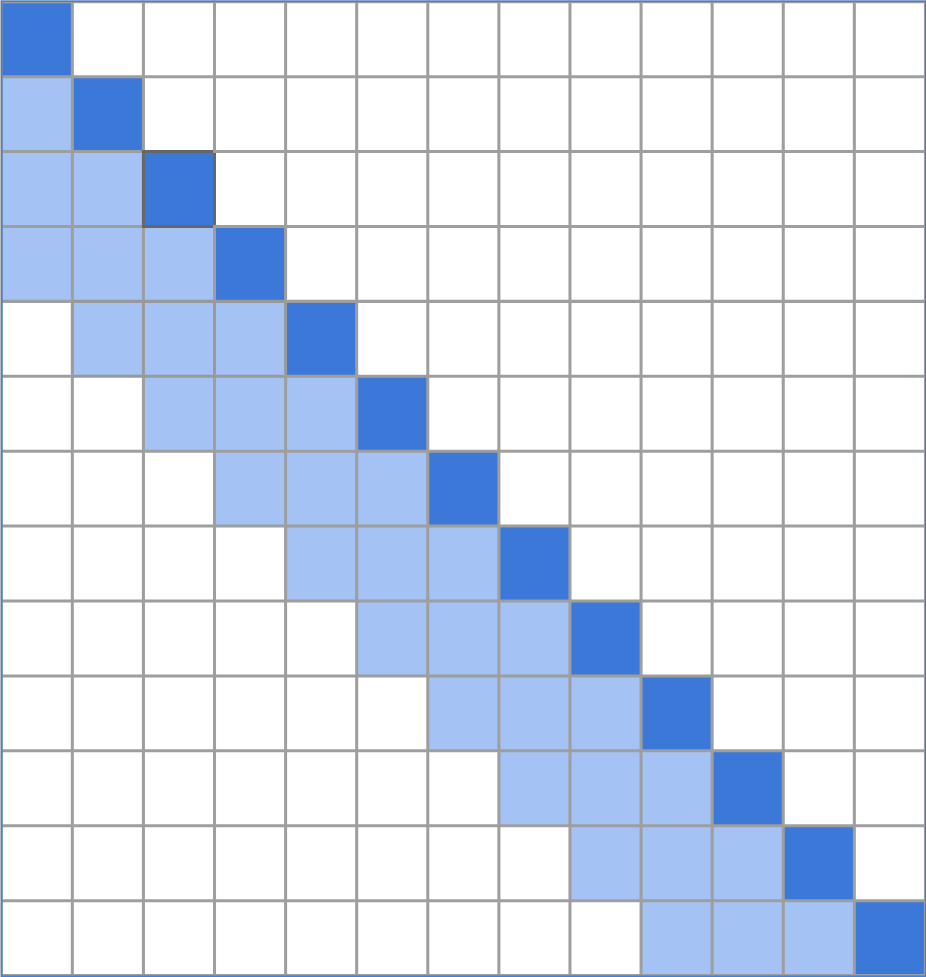}
  \caption{Local attention}
  \label{fig:sfig1}
\end{subfigure}%
\begin{subfigure}{.33\textwidth}
  \centering
  \includegraphics[width=.8\linewidth]{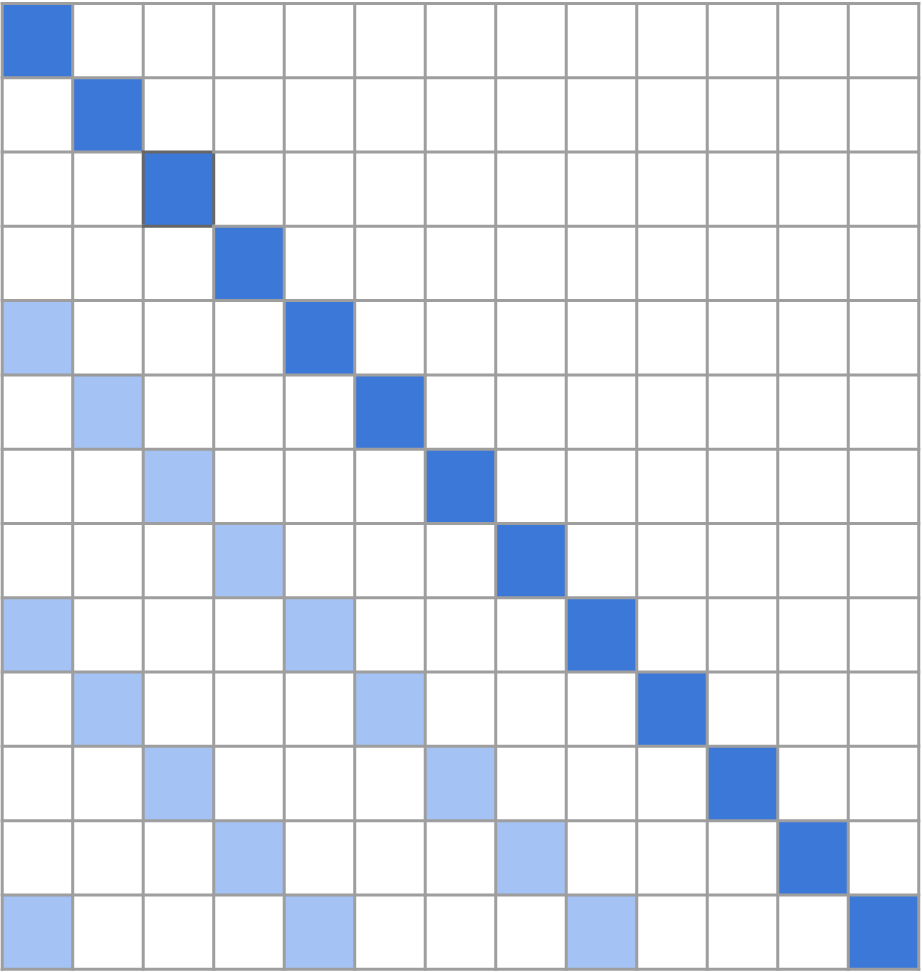}
  \caption{Strided attention}
  \label{fig:sfig2}
\end{subfigure}
\begin{subfigure}{.33\textwidth}
  \centering
  \includegraphics[width=.8\linewidth]{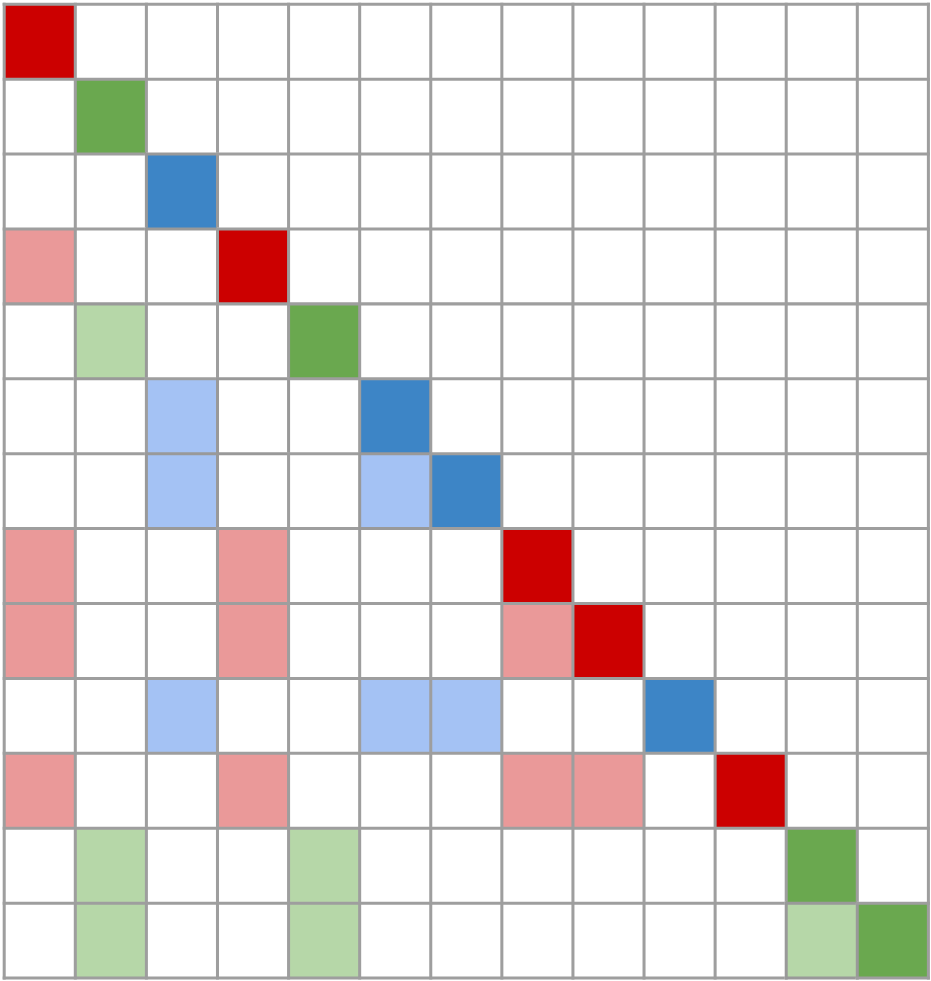}
  \caption{Routing attention}
  \label{fig:sfig3}
\end{subfigure}
\caption{Figures showing 2-D attention schemes for the Routing Transformer compared to local attention and
strided attention of \cite{child2019generating}. The rows represent the outputs while the columns represent
the inputs. For local and strided attention, the colored squares represent the elements every output row
attends to. For attention routed as in Section~\ref{sec:routing}, the different colors represent cluster memberships
for the output token.}
\label{fig:attention}
\end{figure*}

Since, we route attention via spherical \(k\)-means clustering, we dub our model 
\emph{Routing Transformer}. We give a detailed pseudo-code
implementation for the routing attention computation
in Algorithm~\ref{alg:routing}.
A visualization of the attention scheme
and its comparison to local and strided attention is given 
in Figure~\ref{fig:attention}. The computational complexity of this 
variant of sparse attention is \(O(nkd + n^2d/k)\). 
Cluster assignments correspond to the first term, i.e. 
it compares $n$ routing vectors to all $k$ centroids in 
a space of size $d$. Query/key dot products corresponds to 
the second term, i.e. assuming balanced clusters, each
of the $n$ queries is compared to $n/k$ in its cluster
through a dot product of dimension $d$. Therefore the optimal 
choice of \(k\) is \(\sqrt{n}\) as in \cite{child2019generating}, 
thereby reducing overall memory and computational
cost to \(O\left(n^{1.5}d\right)\) instead of \(O(n^2d)\) 
\citep{vaswani2017attention}.

In practice, we apply mini-batch \(k\)-means to train the cluster centroids.
However, in order to infer balanced routing patterns, we define the sets
$S_i$ to be of equal size roughly \(n/k \sim \sqrt{n}\), i.e. 
for every centroid \(\mu_i\) we sort tokens by distance to \(\mu_i\) and cluster 
membership is determined by this threshold (\(\operatorname{top-k}\)). This adds an
additional \(O(n \log{n})\) term to the cost, however note that 
this is eclipsed by the dominating term of \(O(n^{1.5}d)\).
This strategy is simple and efficient. In particular, it guarantees that all clusters
have the same size, which is extremely important in terms of computational efficiency
on parallel hardware like graphic cards. As a downside, this assignment does not
guarantee that each point belongs to a single cluster.
In the future, we want to 
investigate using balanced variants of \(k\)-means \citep{banerjee2004frequency,malinen2014balanced} 
which is not common in an online setting.

During training, we update each cluster centroid \(\mu\) by an exponentially moving average 
of all the keys
and queries assigned to it: 
\begin{align*}
\mu \gets \lambda \mu + \frac{(1 - \lambda)}{2} \sum_{i: \mu(Q_i) = \mu} Q_i + \frac{(1 - \lambda)}{2} \sum_{j: \mu(K_j) = \mu} K_j,
\end{align*}
where \(\lambda\) is a decay parameter which we usually set to \(0.999\).
Additionally, we also exclude padding tokens from affecting the centroids.

\begin{algorithm}
\caption{Routing Attention}\label{alg:routing}
\begin{algorithmic}[1]
\State Queries, Keys and Values: $Q, K, V \in \mathbb{R}^{n \times d}$
\State Centroid: \(\bm{\mu} \in \mathbb{R}^{k \times d}\)
\State decay: \(\lambda\)
\If {\texttt{left to right mask}}
\State $K \gets Q$
\EndIf
\LineComment{Normalize to unit ball}
\State $Q \gets \operatorname{LayerNorm}(Q)$ \Comment{scale, bias disabled}
\State $K \gets \operatorname{LayerNorm}(K)$ \Comment{scale, bias disabled}
\State $Q_{prod} \gets \bm{\mu} Q^\top$ \Comment{$k \times n$}
\If {\texttt{not left to right mask}}
\State $K_{prod} \gets \bm{\mu} K^\top$ \Comment{$k \times n$}
\EndIf
\State \(w \gets n/k\) \Comment{attention window}
\State $Q_{idx} \gets \operatorname{top-k}(Q_{prod}, w) $ \Comment{$k \times w$}
\State $Q_{idx} \gets \operatorname{sort}(Q_{idx})$ \Comment{sort to preserve order}
\State $K_{idx} \gets Q_{idx} $ \Comment{$k \times w$}
\If {\texttt{not left to right mask}}
\State $K_{idx} \gets \operatorname{top-k}(K_{prod}, w) $ \Comment{$k \times w$}
\State $K_{idx} \gets \operatorname{sort}(K_{idx})$ \Comment{sort to preserve order}
\EndIf
\State \(Q' \gets \operatorname{gather}(Q, Q_{idx})\) \Comment{$k \times w \times d$}
\State \(K' \gets \operatorname{gather}(K, K_{idx})\) \Comment{$k \times w \times d$}
\State \(V' \gets \operatorname{gather}(V, K_{idx})\) \Comment{$k \times w \times d$}
\State $A \gets Q' (K')^\top$ \Comment{$k \times w \times w$}
\If {\texttt{left to right mask}}
\State \(A \gets \operatorname{ltr}(A)\)
\EndIf
\State $A \gets \operatorname{softmax}(A)$. \Comment{$k \times w \times w$}
\State \(V' \gets \operatorname{einsum}(kww, kwd \rightarrow kwd, A, V')\)
\State \(X \gets \operatorname{scatter}(K_{idx}, V')\) 
\State $Q_m \gets \operatorname{one-hot}(\arg\max(Q_{prod}))$ \Comment{$k \times n$}
\State $K_m \gets \operatorname{one-hot}(\arg\max(K_{prod}))$ \Comment{$k \times n$}
\LineComment{Update centroids}
\State $\bm{\mu} \gets \lambda \bm{\mu} + (1 - \lambda)Q_mQ/2 + (1 - \lambda)K_mK/2$
\State \Return{$X$}
\end{algorithmic}
\end{algorithm}

There is an additional nuance regarding clustering queries and keys that comes into play
when using causal attention (i.e. left to right masking), as is usually the case in language 
models. When grouping queries and keys belonging to a certain cluster centroid \(\mu\), we 
may get as members queries \(Q_i\) for keys \(K_j\) where time-step \(i \le j\). This therefore requires an additional masking strategy in addition to the lower triangular mask 
used for causal attention. One solution that avoids
having to use an additional mask, is to simply share keys and queries. Empirically, we have found that
this works at par or better than separate keys and queries together with an additional masking strategy
in the causal attention setting. For encoder self attention and encoder-decoder cross-attention, additional masking
or sharing queries and keys is not necessary.

\section{Experiments}\label{sec:experiments}
We evaluate our sparse attention model on various generative modeling tasks
including text and image generation. 
The following sections report our
results on \texttt{CIFAR-10}, \texttt{Wikitext-103} \citep{merity2016pointer},
\texttt{enwik-8} \citep{mahoney2011large}, \texttt{ImageNet-64} 
as well as \texttt{PG-19} \cite{rae2020compressive}. 
We find that a scaled up version of local attention is a surprisingly strong baseline
and that our Routing Transformer outperforms Transformer-XL 
\citep{dai2019transformer} and the Sparse Transformer model of
\citet{child2019generating} on all tasks. 
On the recently released \texttt{PG-19} data-set, 
we find that local attention again is a strong baseline,
with a slightly worse performance compared to Transformer-XL \citep{dai2019transformer}.
We also find that the Routing Transformer model out-performs both Transformer-XL 
\citep{dai2019transformer} and Compressive Transformer \citep{rae2020compressive}, setting a 
new state-of-the-art result. 

In all our models except the one used for \texttt{PG-19}, we allocate
half the heads to do local attention and the other half
to route attention as in Equation~\ref{eq:high-res}.
For all our experiments except for \texttt{PG-19},
we use the Adam optimizer \citep{adam} with learning rate \(2\times 10^{-4}\) with
\(\beta_1 = 0.9\) and \(\beta_2 = 0.98\) following the learning
rate schedule described in \citet{vaswani2017attention}. We train all models on \(128\) TPUv3 cores.
The setup used for \texttt{PG-19} is described in Section~\ref{sec:pg19}.

\subsection{CIFAR-10}\label{sec:cifar}
\texttt{CIFAR-10} is a widely used image data-set which consists of \(60,000\) colored
images of size \(32\times 32\). Since the sequence lengths in this case are relatively
short (\(3072\)), we use this as a toy data-set to perform
various ablations to tease apart the effect of various hyper-parameter choices
on the model performance. We train \(12\) layer models
with a total of \(8\) attention heads, and report a comparison of the effect 
of various hyper-parameter choices on the performance and speed on this data-set. 
In particular, the following hyper-parameters are varied 1) the number of routing attention
heads, 2) the number of routing attention layers and 3) the size of the attention window.
For routing attention we use \(k = 6\) while varying the attention window, to see the
effect on speed and performance. All the \texttt{CIFAR-10} 
models are trained with a batch size of
\(32\) and for a total of \(200,000\) steps. 
In addition, we also compare the Routing Transformer to a \emph{Random Transformer},
where \(K_{idx}\) is randomly chosen rather than being drawn from nearest neighbor search.
For a fair comparison, we take the best model from Table~\ref{tab:cifar10-ablation}
with an attention window of \(512\) and replace all routing heads with \emph{random heads}.
We present the ablation results in Table~\ref{tab:cifar10-ablation} and discuss
it in more detail in Section~\ref{sec:analysis}.
\begin{table*}[h]
\centering
\begin{tabular}{l?c?c?c?c?c}
\toprule
Model &  Routing heads & Routing Layers & Attention window & Bits/dim & Steps/sec \\ 
\midrule
Transformer & 0 & 0 & 3072 & 2.983 & 5.608 \\
Local Transformer  & 0 & 0 & 512 & 3.009 & 9.023\\
Random Transformer  & 4 (random) & 8 (random) & 512 & 3.076 & 5.448 \\
\hline
Routing Transformer & 2 & 2 & 512 & 3.005 & 7.968\\
Routing Transformer & 4 & 2 & 512 & 2.986 & 7.409 \\
Routing Transformer & 8 & 2 & 512 & 2.992 & 6.682 \\
\hline
Routing Transformer & 2 & 4 & 512 & 2.995 & 7.379 \\
Routing Transformer & 4 & 4 & 512 & 2.975 & 6.492 \\
Routing Transformer & 8 & 4 & 512 & 2.991 & 5.385\\
\hline
Routing Transformer & 2 & 8 & 512 & 2.995 & 6.442 \\
Routing Transformer & 4 & 8 & 512 & 2.971 & 5.140 \\
Routing Transformer & 8 & 8 & 512 & 3.190 & 3.897 \\
\hline
Routing Transformer & 2 & 12 & 512 & 2.978 & 5.685 \\
Routing Transformer & 4 & 12 & 512 & 2.994 & 4.349 \\
Routing Transformer & 8 & 12 & 512 & 3.400 & 3.062 \\
\hline
Routing Transformer & 2 & 2 & 1024 & 2.975 & 7.344\\
Routing Transformer & 4 & 2 & 1024 & 2.950 & 6.440 \\
Routing Transformer & 8 & 2 & 1024 & 2.982 & 5.192 \\
\hline
Routing Transformer & 2 & 4 & 1024 & 2.990 & 6.389 \\
Routing Transformer & 4 & 4 & 1024 & 2.958 & 5.112 \\
Routing Transformer & 8 & 4 & 1024 & 3.003 & 3.674 \\
\hline
Routing Transformer & 2 & 8 & 1024 & 2.991 & 5.057 \\
Routing Transformer & 4 & 8 & 1024 & 2.983 & 3.597 \\
Routing Transformer & 8 & 8 & 1024 & 3.131 & 2.329 \\
\hline
Routing Transformer & 2 & 12 & 1024 & 2.973 & 4.151 \\
Routing Transformer & 4 & 12 & 1024 & 3.005 & 2.788 \\
Routing Transformer & 8 & 12 & 1024 & 3.291 & 1.711 \\
\bottomrule
\end{tabular}
\vspace{1mm}
\caption{Ablation studies of the Routing Transformer model on the \texttt{CIFAR-10} data-set.
All the models have a total of \(12\) attention layers and \(8\) heads. Routing layers when 
present are always added at the
top of the model. A Routing Transformer model with less than \(12\) routing attention
layers and less than \(8\) routing heads, has the remaining layers and heads of type
local attention. A Random Transformer model has a random attention head in place of
the routing attention head.
We report the performance in bits/dim on the test set and step times
are reported on a TPUv3.}
\label{tab:cifar10-ablation}
\end{table*}

\subsection{Wikitext-103}
\texttt{Wikitext-103} \citep{merity2016pointer} is a large public benchmark
data-set for testing long term dependencies in word-level language models.
It contains over \(100\) million tokens from 28K articles extracted from 
Wikipedia with an average of 3.6K tokens per article,
which makes it a reference data-set to model long-term textual dependencies.
We train a \(10\) layer Routing Transformer with \(16\) heads using the
relative position encoding of \citet{shaw2018self} and with attention and ReLU
dropout rate of \(0.3\) each. For routing attention as in Section~\ref{sec:routing}
we choose \(k = 16\) and attention window to be \(256\) during both training
and evaluation. We describe our results in Table~\ref{tab:wikitext} and 
compare it to other
recent work on sparse or recurrent attention such as Adaptive Inputs
\citep{baevski2018adaptive} and TransformerXL \citep{dai2019transformer}
as well as a local attention with relative position encoding baseline
\citep{huang2018music}.
We find that local attention is a great inductive bias for sparse attention
and is better than the adaptive methods proposed in 
\citet{baevski2018adaptive, sukhbaatar2019adaptive}.
Moreover, our Routing Transformer model is able to get a test perplexity
of \(15.8\) improving on the 18.3 obtained by TransformerXL
\citep{dai2019transformer} while having fewer self-attention layers, 
and without the need for segment level recurrence.

\subsection{enwik-8}
The \texttt{enwik-8} \citep{mahoney2011large}
is a data-set to benchmark text compression 
algorithms in the context of the Hutter prize. This data-set
consists of the first 100M bytes of unprocessed Wikipedia.
It is typically used to evaluate character-level language models.
Similar to the prior work of \citet{dai2019transformer, child2019generating}
we use a sequence length \(n = 8192\) and benchmark our results
against various baselines including local attention.
We train a \(24\) layer model with \(8\) attention heads with an
attention and ReLU dropout rate of \(0.4\) each and using the relative
position encoding of \citet{shaw2018self}. For routing
attention as in Section~\ref{sec:routing} we set \(k = 32\) 
and attention window \(256\). We report perplexity of $0.99$ like 
TransformerXL and Sparse Transformer, slightly under $0.98$ from 
Adaptive Transformer. 

\subsection{ImageNet \(64\times 64\)}
In order to evaluate the ability of our model to capture long term
dependencies on a modality other than text, we report results on
the \texttt{ImageNet} \(64\times 64\) data-set as used in \citet{child2019generating}.
For auto-regressive image generation, this data-set consists of images of 
$64\times 64 \times 3$ bytes represented as long sequences of length $12,288$ 
presented in raster scan, red-green-blue order.
We train a \(24\) layer model with \(16\) attention heads, with half
the heads performing local attention, and the other half routing
attention as in Section~\ref{sec:attention}. For routing attention
we set \(k = 8\), attention window \(2048\), batch size \(1\) 
and train our model for roughly \(70\) epochs as in 
\citet{child2019generating}. We compare
our model to a scaled-up ImageTransformer model with local attention \citep{parmar2018image}
and the SparseTransformer model of \citet{child2019generating}.

We find that local attention \citep{parmar2018image} is a 
strong baseline for image generation, obtaining
\(3.48\) bits/dim when scaled up to \(24\) layers and \(16\) heads, compared
to later work like Sub-scale Pixel Networks (SPN) \citep{menick2018generating}.
Our Routing Transformer model achieves a performance of \(3.425\) bits/dim 
(see Table~\ref{tab:imagenet}) compared
to the previous state-of-the-art of \(3.437\) bits/dim \citep{child2019generating}, 
thereby showing the advantage of the content based
sparsity formulation of Section~\ref{sec:routing}.
\begin{table*}[h]
\centering
\begin{tabular}{l?c?c?c}
\toprule
{Model} & Layers & Heads & Perplexity \\ 
\midrule
LSTMs \citep{grave2016improving} & - & - & 40.8 \\
QRNNs \citep{merity2018analysis} & - & - & 33.0 \\
Adaptive Transformer \citep{sukhbaatar2019adaptive} & 36 & 8 & 20.6 \\
Local Transformer & 16 & 16 & 19.8 \\
Adaptive Input \citep{baevski2018adaptive} & 16 & 16 & 18.7 \\ 
TransformerXL \citep{dai2019transformer} & 18 & 16 & 18.3 \\
\hline
\midrule
\emph{Routing Transformer} & 10 & 16 & \textbf{15.8}\\
\bottomrule
\end{tabular}
\vspace{1mm}
\caption{Results on language modeling on \texttt{Wikitext-103} data-set. 
Local Transformer refers to Transformer \citep{vaswani2017attention} with relative
position encoding \citep{shaw2018self} together with local attention. 
Perplexity
is reported on the test set.}
\label{tab:wikitext}
\end{table*}

\begin{table*}[h]
\centering
\begin{tabular}{l?c?c?c}
\toprule
{Model} & Layers & Heads & Bits per byte \\ 
\midrule
T64 \citep{al2019character} & 64 & 2 & 1.13\\
Local Transformer & 24 & 8 & 1.10 \\
TransformerXL \citep{dai2019transformer} & 24 & 8 & 0.99  \\
Sparse Transformer \citep{child2019generating} & 30 & 8 & 0.99 \\
Adaptive Transformer \citep{sukhbaatar2019adaptive} & 24 & 8 & \textbf{0.98} \\
\hline
\midrule
\emph{Routing Transformer} & 12 & 8 &  0.99 \\
\bottomrule
\end{tabular}
\vspace{1mm}
\caption{Results on language modeling on \texttt{enwik-8} data-set.
Local Transformer refers to Transformer \citep{vaswani2017attention} with relative
position encoding \citep{shaw2018self} together with local attention.
Bits per byte (bpc) is reported on the test set.}
\label{tab:enwik}
\end{table*}

\begin{table*}[h]
\centering
\begin{tabular}{l?c?c?c}
\toprule
{Model} & Layers & Heads & Bits/dim \\ 
\midrule
Glow \citep{kingma2018glow} & - & - & 3.81 \\
PixelCNN \citep{van2016conditional} & - & - & 3.57 \\
PixelSNAIL \citep{chen2018pixelsnail} & - & -& 3.52 \\
SPN \citep{menick2018generating} & - & - & 3.52 \\
ImageTransformer \citep{parmar2018image} & 24 & 16 & 3.48 \\
Sparse Transformer \citep{child2019generating} & 48 & 16 & 3.44 \\
Reformer \citep{kitaev2020reformer} & - & - & 3.65 \\
\hline
\midrule
\emph{Routing Transformer} & 24 & 16 & \textbf{3.43}\\
\bottomrule
\end{tabular}
\vspace{1mm}
\caption{Results on image generation on \texttt{ImageNet- 64} in bits/dim.}
\label{tab:imagenet}
\end{table*}

\subsection{PG-19}\label{sec:pg19}
\texttt{PG-19} is a new data-set released by \citet{rae2020compressive}
which is larger and longer than previous language modeling data-sets.
The data-set is created from approximately \(28,000\) Project Gutenberg books published before 
1919, consisting of \(1.9\) billion tokens 
and comprises an average context size of roughly \(69,000\) words. This is text that is 
\(10\times\) longer in context than all prior data-sets such as \texttt{Wikitext-103}, with minimal pre-processing
and an open vocabulary that makes it extremely challenging for long text modeling tasks.
We use a subword vocabulary of size approximately 98,000 and report perplexities normalized
by the token counts reported in \citet{rae2020compressive}.
On this data-set we train a \(22\) layer Routing Transformer model with \(8\) heads with a 
sequence length of \(8192\)
and set a new state-of-the-art result on this data-set, improving on both Compressive 
Transformers 
\citep{rae2020compressive}, as well as Transformer-XL \citep{dai2019transformer}.
For this data-set we change our training setup in three ways. Firstly, we use only \(2\) routing heads instead of 
sharing it equally with local heads. Secondly, we use routing heads only in the last two layers of the model instead
of having them present in every layer. This is motivated by our empirical finding that long range attention is 
only needed in the last few layers - see also \citet{rae-razavi-2020-transformers}. Finally, we use
the Adafactor optimizer \citep{shazeer2018adafactor} which is more memory efficient than Adam 
in training larger models. We use a learning rate constant of \(0.01\) with a linear warmup over
\(10,000\) steps followed by a \emph{rsqrt_normalized_decay}. We do not make use of any dropout, or weight decay.
The hidden dimension of our model is \(1032\) and the batch size is \(8192\) tokens. 

From Table~\ref{tab:pg19}, we see that Local Transformer again sets a very strong baseline, with a \(24\)-layer
local attention model obtaining a test set perplexity of \(39.3\), while a \(36\)-layer Transformer-XL gets
\(36.3\). Moreover, a \(22\)-layer Routing Transformer model improves on the \(36\)-layer Compressive
Transformer, obtaining a test set perplexity of \(33.2\) compared to \(33.6\), 
while being able to generate sequences of length \(8192\).

\begin{table*}[h]
\centering
\begin{tabular}{l?c?c?c}
\toprule
{Model} & Layers & Heads & Perplexity \\ 
\midrule
Local Transformer & 24 & 8 & 39.3 \\
TransformerXL \citep{dai2019transformer} & 36 & - & 36.3 \\
Compressive Transformer \citep{rae2020compressive} & 36 & - & 33.6 \\
\hline
\midrule
\emph{Routing Transformer} & 22 & 8 & \textbf{33.2}\\
\bottomrule
\end{tabular}
\vspace{1mm}
\caption{Results on language modeling on \texttt{PG-19} data-set. 
Local Transformer refers to Transformer \citep{vaswani2017attention} with relative
position encoding \citep{shaw2018self} together with local attention. 
Perplexity is normalized by the number of tokens reported in \cite{rae2020compressive}
and is reported on the test set.}
\label{tab:pg19}
\end{table*}

\section{Analysis}\label{sec:analysis}
\subsection{Local vs Global}
As reported in Section~\ref{sec:experiments}, a scaled up version of 
local attention is a strong baseline for efficient attention over long sequences. 
From Table~\ref{tab:cifar10-ablation} we see that local
attention is slightly worse than full attention - \(3.009\) vs \(2.983\) bits per dim. 
Adding \(2\) routing layers with \(4\) heads almost closes the gap
with the performance of full attention, achieving \(2.986\) bits per dim. 
Adding more routing layers and heads 
improves performance up to a point, with the best performing model
with an attention window of \(512\) having \(4\) routing layers and \(4\) routing heads,
and achieving \(2.975\) bits per dim.
Increasing the attention window from \(512\) to \(1024\) uniformly results in improvement
in every setting. 
The best model on \texttt{CIFAR-10} has an attention window of \(1024\) with
\(4\) routing layers and \(4\) routing heads.
Interestingly, the best Routing Transformer models perform better than full 
attention, but not by a large enough amount to rule out noise. 
More importantly,  Table~\ref{tab:cifar10-ablation} shows
the importance of local attention in building intermediate representations, 
with a model with only routing attention layers and heads with attention windows of
\(512\) and \(1024\) achieving $3.400$ and $3.291$ bits per dim respectively.

Thus Table~\ref{tab:cifar10-ablation} shows us the importance of local representations,
as well as the benefit of adding a few routing layers and heads to enforce
a more global representation. Since attention weights are a probability distribution on
the entire set of tokens, we evaluate the difference in attention patterns between local and
routing attention by computing the Jensen-Shannon divergence between 
the two kinds of attention distributions for a random subset of heads 
in our network on the \texttt{Wikitext-103} data-set.
The divergence is computed over the entire sequence length of \(4096\).
We average over \(10\) runs and report means and standard deviations of the \(\operatorname{JSD}\) in Table~\ref{tab:jsd}. 
Note that the \(\operatorname{JSD}\) is always non-negative and is upper-bounded by \(0.6931\)
when computed using the natural logarithm.
We observe that the divergence between the different
local heads is always very low compared
to the divergence between local and routing attention heads, 
which is almost always very close to the upper-bound
of \(0.6931\). 
Divergence between different routing attention heads falls somewhere in
between, being closer to the upper-bound.
This shows that the attention distribution inferred 
by the routing attention of Section~\ref{sec:routing}
is highly non-local in nature and 
different heads specialize in attending to very different parts 
of the input.

\begin{table*}[h]
\centering
\begin{tabular}{l?c?c?c}
\toprule
 & \(\operatorname{JSD}(local \lVert local)\) &  \(\operatorname{JSD}(local \lVert routing)\) & \(\operatorname{JSD}(routing \lVert routing)\) \\ 
\midrule
\texttt{layer 0} & \(0.0038 \pm 0.0018\) & \(0.4706 \pm 0.0319\) & \(0.1579 \pm 0.0576\) \\
\texttt{layer 1} & \(0.3071 \pm 0.1217\) & \(0.6674 \pm 0.0153\) & \(0.5820 \pm 0.0104\) \\
\texttt{layer 2} & \(0.2164 \pm 0.0803\) & \(0.5896 \pm 0.0249\) & \(0.4015 \pm 0.0121\) \\
\texttt{layer 3} & \(0.1163 \pm 0.0336\) & \(0.6047 \pm 0.0181\) & \(0.4144 \pm 0.0264\) \\
\texttt{layer 4} & \(0.1840 \pm 0.0562\) & \(0.6266 \pm  0.0062\) & \(0.4191 \pm 0.0879\) \\
\texttt{layer 5} & \(0.2284 \pm 0.0225\) & \(0.6463 \pm 0.0155\) & \(0.4687 \pm 0.0449\) \\
\texttt{layer 6} & \(0.1901 \pm 0.0525\) & \(0.6471 \pm 0.0040\) & \(0.5175 \pm 0.0469\) \\
\texttt{layer 7} & \(0.1566 \pm 0.0685\) & \(0.5798 \pm 0.0235\) & \(0.4350 \pm 0.0139\) \\
\texttt{layer 8} & \(0.1638 \pm 0.0739\) & \(0.5993 \pm 0.0148\) & \(0.4268 \pm 0.0291\) \\
\texttt{layer 9} & \(0.2095 \pm 0.0560\) & \(0.6127 \pm 0.0053\) & \(0.3581 \pm 0.0019\)\\
\bottomrule
\end{tabular}
\vspace{1mm}
\caption{Jensen-Shannon divergence between 
the attention distributions of a random
local attention head and a random head that routes attention 
as in Section~\ref{sec:routing}
per layer on the \texttt{Wikitext-103} data-set. 
We report means and standard deviations computed over \(10\) runs and 
use the natural logarithm so that divergences are upper-bounded by 
\(0.6931\).}
\label{tab:jsd}
\end{table*}

Qualitatively, from the ablations in Table~\ref{tab:cifar10-ablation},
we hypothesize that the reason for the strong performance of the Routing Transformer
is due to the fact that it combines building local representations over several
layers, together with enforcing global consistency for every token. This is achieved
via an approximate Maximum Inner Product Search (MIPS) over the entire set of tokens
(see Section~\ref{sec:routing}),
and selecting pairs that have a high dot product for attention.
This allows various entities such as gender, nouns, dates and 
names of places to be consistent throughout the entire sequence, 
since on expectation the dot product similarity
between similar entities are high, while for differing entities they are expected to be low. 
Essentially, we conjecture that for every time step, the prediction depends on a small 
support of \emph{high value} tokens:
local attention facilitates local consistency and fluency, while a full dot product attention
would facilitate global consistency. However, for long sequences since full attention is 
infeasible, we believe that using spherical \(k\)-means to perform a MIPS search over the 
global set of tokens and performing attention between these high dot product items
is a good approximation to \emph{full dot product attention}. The importance of the MIPS
search to select high dot product items is highlighted from the ablation in 
Table~\ref{tab:cifar10-ablation}, where we see that a Random Transformer 
performs worse compared to a Local Transformer and a Routing Transformer
with the same configuration, (\(3.076\) vs \(3.009\) vs \(2.971\)) bits/dim.

\subsection{Recurrence vs Sparse Attention}
We also note that sparse attention is an orthogonal approach to that of Transformer-XL and 
Compressive Transformer,
which train on small sequences and by performing careful cross attention over cached previous chunks hope to generalize to longer sequences. By contrast, we directly train on long sequences from the beginning - 
e.g., the Compressive Transformer trains on chunks of size \(512\) for \texttt{PG-19}, while 
we train on sequences of length \(8192\). 
The benefit of the Transformer-XL like approach is that it is less memory consuming
and thus is able to scale to \(36\) layers. 
Sparse attention (including local attention) on the other hand is more memory 
expensive since it trains directly on long sequences and therefore
can scale to fewer layers for the same problem. However, as we demonstrate, it is competitive
with the Transformer-XL like approaches even when using fewer layers and is guaranteed to generalize to the long 
sequence length that it was trained on.

\begin{table*}[h]
\centering
\begin{tabular}{l?c?c?c?c?c?c}
\toprule
{Model} & Dataset & Seq. length & Layers & Heads & Attention window & Steps/sec \\ 
\midrule
Local Transformer & \texttt{PG-19} & 8192 & 24 & 8 & 512 & 1.231\\
\emph{Routing Transformer} & \texttt{PG-19} & 8192 & 22 & 8 & 512 & 0.7236\\
\bottomrule
\end{tabular}
\vspace{1mm}
\caption{Step time comparison between Local Transformer and 
Routing Transformer on a TPUv3 for the \texttt{PG-19} 
data-set.}
\label{tab:step-time}
\end{table*}
\subsection{Wall-clock time}
We compare the step times for training the various sparse attention models on
the \texttt{CIFAR-10} data-set in Table~\ref{tab:cifar10-ablation} as well as
on the \texttt{PG-19} data-set in Table~\ref{tab:step-time}. For \texttt{PG-19}
we report only a comparison between the Local Transformer and the Routing Transformer,
since sequence lengths are \(8192\) and performing full attention is infeasible.
All the step time comparisons are made on a TPUv3, with the same number of cores and
batch sizes to facilitate a fair comparison.
As we see from Table~\ref{tab:cifar10-ablation} local attention is much faster than
full attention, training at \(9.023\) steps per second
compared to \(5.608\) steps per second. The Routing Transformer models on
\texttt{CIFAR-10} have step times that depend on the number of routing heads,
with the best performing model with the same attention budget as local attention (i.e.
an attention window of \(512\)), which has \(8\) routing layers and \(4\) routing heads,
training at \(5.140\) steps per second. Other Routing Transformer models are faster
while still matching full attention, e.g.,  \(2\) routing layers with \(4\) routing
heads trains at \(7.409\) steps per second. Therefore, Local Transformer is roughly
between \(1.22 - 1.76\times\) faster than the best performing Routing Transformers.
On the other hand Transformer is between \(0.76 - 1.09\times\) faster than the best
Routing Transformers.

On \texttt{PG-19}, we see from Table~\ref{tab:step-time}, that the Local Transformer
is roughly \(1.7\times\) faster compared to the Routing Transformer, similar to
the trend on \texttt{CIFAR-10}. 
This trade-off with respect to speed compared to the Local Transformer
is due to the lack of support 
for sparse operations on the TPU; on the GPU various sparse kernels
have been proposed which promise to significantly speed up training of these models 
\citep{gale2020sparse}.
Note that our goal in this work is a memory efficient version of sparse attention that can 
well approximate
full attention for long sequences - wall-clock time efficiency is only a secondary goal. 


\section{Conclusion}
Transformer models constitutes the state-of-the-art in auto-regressive generative models
for sequential data. Their space-time complexity is however quadratic in sequence length, due to their attention modules. Our work proposes a sparse 
attention model, 
the Routing Transformer. It relies on content-based sparse attention motivated by non-negative matrix factorization. Compared with local attention 
models, it does not 
require fixed attention patterns but enjoys similar space-time complexity. In contrast with prior work on content-based sparse attention, it does not 
require computing 
a full attention matrix but still selects sparsity patterns based on content similarity.

Our experiments over text and image generation draw two main conclusions. First, we show that a scaled up version of local attention establishes a 
strong baseline on  
modern benchmark, even compared to recent state-of-the-art models. Second, we show that the 
Routing Transformer redefines 
the state-of-the-art in large long sequence
benchmarks of
\texttt{Wikitext-103}, \texttt{PG-19} and \texttt{ImageNet-64}, while being very close to do so on 
\texttt{enwik-8} as well.
Our analysis also shows that routing attention modules offer complementary attention patterns when compared to local attention.

Overall, our work contributes an efficient attention mechanism that 
applies to the modeling of long sequences and redefines the state of the art for
auto-regressive generative modeling. Our approach could prove useful in domains where the inputs are naturally
sparse, such as 3D point clouds, social 
networks, or protein interactions. 

\section{Acknowledgments}
The authors would like to thank Phillip Wang and Aran Komatsuzaki for a Pytorch implementation of Routing Transformer. The authors would also like to thank Yonghui Wu, Weikang Zhou and 
Dehao Chen for helpful feedback in improving the implementation of this work. The authors 
would also like to thank anonymous reviewers and the Action Editor of TACL 
for their constructive comments which helped improve the exposition of this work. 

\bibliography{deeplearn}
\bibliographystyle{acl_natbib}
\appendix

\section{Samples from Routing Transformer}\label{sec:sample}
In the following sections we present a few samples generated from the Routing Transformer trained
on the \texttt{PG-19} data-set with sequence length \(8192\). We use nucleus sampling
\citep{Holtzman2020The} with \(p = 0.8\) and temperature of \(1.0\) to generate these samples. 

\subsection{Sample - I}
During the early days of the Council of Nice and the subsequent existence of the Council of Basle, one section of the Council of Nice made a marked opposition to the pretensions of the Council of Basle. Some of them were men of high rank, others members of the lower classes. They had been formed into a union which was called the Papal Council, and which for the time being proved to be of the utmost importance to that Church in which it met. The first session of the Council of Nice took place on September 2, 1487. The two bodies met in solemn assembly and made arrangements with one another. It was decided that a considerable portion of the Council should proceed to Zurich and lay before the Council of Constance the proposals of the Church of Basle for settling their disputes. It was also resolved that a meeting of the representatives of the Christian Emperors of Germany, France, Portugal, Hungary, England and France should be held. Three bishops were commissioned to despatch ambassadors to each of the two Councils to urge their respective envoys to combine and come to some agreement regarding matters ecclesiastical. It was agreed that the Council of Basle should at once take steps for the reformation of the Church and the peace of Christendom; while the two meetings were to be united in one. Various questions of dispute were settled in a friendly way; but the whole subject of the relations of the Church to the Papacy was laid before the Council of Basel, and an agreement arrived at regarding the ecclesiastical and civil relations of the Church with the head of the Papacy.One important result of this Council was that it thus obtained two important concessions from the Popes: the first in making a papal establishment the natural basis of ecclesiastical authority on a great scale and yielding to the papal pretensions; and the second in providing for a Papal Council of Basle in which there should be ecclesiastical authority, and a bishop of the Roman Church, to meet the needs of the Churches of Europe. The Council of Basle likewise obtained the provision that the election of the Pope should be conducted by the same general council and by the head of the Church at Rome, and that no other form of appointment than that of a personal election to the Papacy should be in force. It was in effect a completion of the Council of Basle. It left without a head, indeed, but with an indication of its existence, the crowning work of the nineteenth century. The Council of Basle had not succeeded in bringing about the acceptance of the Papal headship; but there can be no question that the defeat of the Papal claim, at the Council of Lyons in the year following (December 17,1530), determined the attitude of the Papacy towards the Church, and prepared the way for the action of the Council of Trent. For at that time it seemed as though, after the Council of Lyons, the Council of Trent could no longer prevent the intrusion of the Papacy into the Church, and it was recognised that there was to be no more preaching in the Churches of Europe, for this once. Yet the fact remains that there was no Papal interference with Church government. From that time forward,however, the rule of the Church became more rigorous, and towards the end of the sixteenth century began the crisis in the Church which lasted until the general council of the Council of Trent.The organisation of the Swiss Church had been brought down to the time of Zwingli (1516-1531). It was based upon an organisation strictly clerical in character, as the Canons of the Roman Church insisted upon the clergy being for the most part clerics of the clerical order. In this respect this system was a reminiscence of that of the Roman Church, except that the mass of the people were clerics of the clerical order, who were liable to be deposed at any moment by the spiritual authorities. In the present instance we must recognise that Zwingli introduced a new conception of Church government; for although a great deal of the work of the Reformation was done under the direction of Zwingli, yet the organisation of the Swiss Church to some extent, and the connection of the civil with the ecclesiastical system, served as models for the organisation of the Church in all the Protestant lands.No doubt there was a great amount of copying of Rome, and some irregularities of arrangement were to be found. It is to be noted, however,that most of the reformation principles and practices of the Reformation were embodied in the Church organisation of the Swiss Protestants; the chief result being that, whereas the earlier system was still simple, the Church reformed more strongly and specifically, and was thereby destined to get more help in the direction of the Protestant reformation. So that even in the confusion arising from the change of Church organisation in the sixteenth century the Swiss Church was drawn much more closely to Rome than it would otherwise have been.The first work of the Reformation, however, to which the introduction of the Bible is to be attributed, was done in the early years of the sixteenth century. The era of the Reformation had begun; and this event was by no means likely to pass over without some indication of its influence in the world, for the Reformation had assumed the character of a great political event. The work of the reformation was in a large degree concerned with the national character of Protestantism. The reformation had been the work of religious philosophers, and it was a momentous and noteworthy step towards the winning of the political independence of the nations. But Luther had accomplished no permanent political revolution.Instead of that he had worked to establish that political absolutism of the kings which is the most distinctive characteristic of the Protestant polity. It was not in the modern European sense that he destroyed feudalism and other institutions based on tradition; for the victory was of the Gospel, and he hoped by its means to add another to the ten thousand proofs of the Divine origin of the kingdom of God. The power which he had created was in a large sense political power, and it was part of his function to secure such political power for the Church. He also worked, at an early stage, to further the establishment of the independence of the Church of the Brethren, but it was not until the Reformation became an aggressive factor in the life of the nation that the need for further political recognition of the Church was felt.The reformation movement was to have a most important effect on other aspects of the life of the people, and also upon the growth and extension of Protestantism. The great change which was thus produced, and which has been described as the direct and immediate outcome of the Reformation, was in effect essentially religious in its nature. The re-establishment of the Church of the Brethren has never been one of the least noteworthy phases in the history of the nation. During the next two centuries the popular Church of the Brethren increased in number, importance, and popularity. The king, the nobles, and the more educated portion of the people came more and more to regard it as the natural bulwark of Protestantism; and in a comparatively short time, and within comparatively short space of time, that work which Luther did for the establishment of the national life has been carried to a high degree of accomplishment by the English and other Protestant communities. The beginning of the Reformation, as already indicated, was a direct consequence of the effects which were brought about by the Reformation.It was not simply in the Church that the recognition of the Church of the Brethren made itself felt. The religious feelings which were aroused,and which were finally developed into a religious habit, have already been sufficiently dealt with in connection with the general history of the German nation; and the re-establishment of a purely spiritual faith and of a dominant religious life in the land, one which could not possibly have been attained save by the outpouring of the Holy Spirit and by the renewing and transforming influences of the Divine Spirit, was among the first results of the Reformation. To that work belongs the development of the German Reformation in its broadest and widest form; and the causes which determined the course of that development may be shortly stated as follows:In the first place, we have seen how the study of the Bible and of the Apocrypha, and of the Jewish conception of God and of the obligation to fast, created a desire for the study of the Bible in a larger and deeper manner than any before known; and, secondly, how the passion for writing profane history and for the writing of sacred history was fostered by the increase of the Roman Church; and,thirdly, how the study of the Scripture in a more liberal spirit--a great impulse to the study of the Old Testament in an earlier period in all its forms, and towards a development of the conception of God and of a more secular spirit in the life of the nation, helped to accelerate the spread of a new and healthier conception of the Christian life. This latter result, and this alone, tended to produce a new and productive condition of the nation in the matter of religion; but it also reacted on the missionary endeavours of the members of the Church of the Brethren to attain a deeper religious development. The desire to read the Bible, to adopt the principles of the Reformers, and to raise the standard of life and manners, not only stimulated the energy and assisted the zeal of the societies of the Brethren, but also stimulated their wider application to particular branches of the work which they had to do. In other words, the deeper study of the Bible as the study of the Old Testament became the religion of the people, and by the sheer force of the influence of these early studies the religious work of the German Reformation took shape, and became one of the most important political movements in Germany. The movement,thus inaugurated, was still later in reaching results in other countries.Before reaching Germany, however, the religious work of the Reformation had made a great impression upon one of the rulers of that country. Philip of Hesse, in 1495, was a child in years; but he was a man of religious instincts and aspirations, and his first utterances were destined to be the embodiment of that new religious idea which for so many years had been deeply implanted in the national mind. The importance of this movement will not be denied. It was an expression of the revival of the primitive and devout tendencies in the Lutheran Church; and in the Lutheran Reformation itself there was far less of scientific study than of poetic expression. It is plain, then, that the Reformation movement in Germany was in some respects influenced, as it was also in some respects modified, by the study of the Scriptures in their original languages, and with a more modern translation of the Bible into modern German.But the condition in which the Reformation found Germany had in a large measure changed. The enthusiasm which formerly animated men for the study of the Bible in all its original tongues was broken down. They did not recognize that the Bible for the understanding of God's Word, and for its guidance through life, was not only the best language in which it was written, but, as already noticed, it was the chief interpreter of all other languages. When we remember that Luther was a professor of Divinity at Wittenberg; that Luther had expounded, in the German tongue, his new faith and new life; and that this same translation had found its way into the minds of thousands and tens of thousands of people in other countries; and that the old German Bibles did not by any means constitute the translation generally used, and that, except for the selection of modern translations, the standard text of the German Bibles for our service was of course by no means the best, we can hardly fail to see that it became clear that the Scriptures as the Bible for the understanding of God's Word were inadequate for the elucidation of religious problems; and, further, that there was no substitute, no adequate translation of the Bible that was available.In order that the question of its complete translation might be understood, it was necessary to seek to adapt it to the spirit and needs of Germany, and this was the task which the Government of the German Empire set itself, and upon the result of which depended the situation under which the Reformation came about.The aim of the Reformation, in the words of Luther himself, was,primarily, the study of the Bible as a living interpreter of God's words and revealing God's will in them; and, secondarily, the acquisition of a living, active, self-interpreting, and God-glorifying Christian spirit. In order to study the Old Testament as a living revelation of God's character and as an example of what God's Spirit, as that Spirit of truth, is capable of doing, it was essential that they should have some historical contact with the Old Testament; and this contact was brought about by the introduction of commentaries on its text. It was in this way that the institution of the /Kleinpostille/ and the growth of a literature for it were due to the zealous and devoted efforts of German Christians at this period. It was because the /Kleinpostille/ and the/Kleinpostille-Lexicon/ were due to the vigorous, self-denying, energetic,and helpful German literature which sprang up in Germany during this period, that the celebrated /Lutherana/ was put forth in the sixteenth century.Nor did it remain for the Reformation to avail itself of the facilities which this literary form gave it in Germany. It had not been intended to continue its work without the aid of a translation, and before it was generally accepted as such an adequate one, a work of translation had to be done, and this was accomplished in a most able and painstaking fashion by /The Commentary on the Galatians/ in 1531. In that work, also, the advantages of translation, as well as the emphasis which the services which it rendered were warranted to lay upon, were well recognised, and it has always been thought that Luther's translation was the best rendering that was available for his readers.There is no need to dwell upon the fact that a work such as this,which for twenty years was in the hands of all the students of German theology, could not have found its way to a Christian home in a Protestant country like Germany without being a source of new and most valuable information. We find, indeed, in it the most valuable reflection on the extent of the religious life and the condition of culture in the countries which represented the belief and received the teachings of the Reformation, as well as the most remarkable revelation of the kind which the Lutheran Reformation contain\\

\subsection{Sample - II}
which the king and his council had agreed upon. On Sunday morning at eleven o'clock I arrived at the royal palace of Paris, where my uncle,the bishop of Chartres, received me in the grand antechamber with the customary grace of his manner. We went immediately into the room of the king, and the bishop of Chartres was so kind as to take me to him in the presence of his majesty. This morning Louis XVIII. held a review of the troops under the orders of the Duke of Orleans.[Illustration: THE OLD CHATEAU (ST. GERVAIS)]"I did not observe," adds the Abbé de Pradt, "that he had a very fine set of teeth, although it is not the custom in the court of France. I was struck by the extreme whiteness of his countenance, and the whiteness of the beard, which he allowed me to see and feel. He was still very pale, and his clothes alone gave him the appearance of being in good health." He spoke to me in a low and gentle tone without any affectation of severity.[Illustration: LOUIS-PHILIPPE DE FRANCE, SON OF LOUIS XVIII. AND CHARLOTTE CORDAY.]He was tall, but looked thin; his frame was very lean, and he did not possess sufficient dignity to conceal the feebleness arising from the length of his limbs and the length of his legs. He walked like a man who is too proud, and who does not wish people to see him. All those who had the honor of being admitted to the royal bedchamber immediately remarked his extreme nervousness. This state of the King's character, which has been much remarked, arises from the long period of preparation for the functions which it occupies, from the long life for which he has been obliged to prepare, and from the weakness of his health. It was natural that the king should not bear arms with all the agility which might be looked for from so young a man. As, however, there was no longer any necessity to employ his bodily strength, he resigned himself to taking a seat, and there he remained motionless for some moments after he had seated himself on a fauteuil. He seemed lost in thought, and his mind must have been deeply occupied. He spoke little. He frequently turned his head to look toward the door; but he did so so slowly that it was impossible to observe his features. At first he showed no interest in the proceedings of the day. At last, a cannon-shot being heard in the direction of St. Cloud,he raised his head, looked for a moment at his watch, and said, "Come now, here is the beginning of the play." I afterwards saw him every day in the same manner, and the habit of not looking for the end of the piece continued in his mind until his death.It was only in some moments of extreme agitation or deep reverie that an expression could be observed upon the King's countenance. His features did not then wear that state of tension which they assumed on the first appearance of serious danger. He did not appear to feel the smallest uneasiness, but, on the contrary, a sort of inward joy.He was full of an instinctive respect for his son's life, and of an anxiety for any danger threatening it. His great anxiety arose from his own extreme weakness as well as from his own inexperience in affairs of state. He was the dupe of his ministers; he regarded them as his real friends and as the most devoted subjects in the world; he would even not deny them the honors he paid to them. He was not disposed, even during his most active occupations, not to forget to send his minister on an important mission.If the King had been a man of energy he would have made active use of his power; but it was a peculiarity which might be said to belong to his whole history to allow himself to be led by others; never to have a will of his own; never to have the courage of his age.The King was very fond of his daughter-in-law, the Princess Louise,Madame Adelaide's only daughter. He was fond also of his daughters.Hortense especially, whom he loved sincerely, was extremely attached to him, and never quitted him without having her clothes pulled, and being told that her petticoats would fall off, in order, she said, that she might walk upon them, as she had never yet worn one. This affection of the poor King for his daughters was so great as to be almost an affection of paternity, and he appeared to be even more attached to them than they to him. The Princess Adelaide, who was also extremely gracious to him, often went with the King the same way; for her great tenderness for her father-in-law, and her own natural timidity, prevented her from ever daring to speak to him upon any political question. The Princesses, though very young, had some influence with their father, the King. Every one would have thought that the Princess Louise had been his wife, and that her father would have been entirely ruled by her wishes, and that this influence would have been an authority upon which he would not have ventured to act; and yet, since his daughter had taken the veil, and had abandoned the Regency, they had seen him frequently on these subjects, and the Princess Louise had been always his companion on the most interesting occasions.When our troops were about Versailles on the 16th of April, 1815, they were fired upon by the Prussian soldiers. The latter had been stationed some hundred paces in the rear of the King's troops, with the object of watching their movements. Suddenly all was changed, or, at least,a sudden silence ensued. At the turn of a road which runs from St. Joseph's chapel to the King's house there was a barricade.The insurgents halted and took up their arms for an instant. The insurgents were very numerous, and had a small but regular force. One of the generals sent forward a soldier to beg permission to fire a few muskets for the purpose of driving back the enemy. The officer advanced to the barricade alone, and returned in about five minutes accompanied by twenty-seven men, all in uniform. They were told to sit down in a circle, and not to stir. Then a man of the people spoke, asking permission to address a word to the general.The people were evidently frightened at this new sort of attack, and were evidently preparing to be frightened. The general, however,continued his calm and dignified demeanor, and began to speak a few words to the people.[Illustration: THE KING'S AMBASSADOR AT THE BARRICADE--Page 58.]"My friends," said he, "I am not surprised to find you ready to give us a demonstration of your love. We need it in our work of salvation, as you need liberty in your work of vengeance. I am about to begin."A man from a group of some thirty men placed himself on the barricade,from a desire to see what was going on. He then cried out loudly:"Forward, forward, my people! Forward!"The King advanced to this barrier. An officer of the national guard stepped forward, and, presenting his musket at their guns, said:"Down with the traitors!" The whole battalion instantly obeyed the order. They were taken, shot, and dispersed, while the royal troops marched along with their muskets at ease, and without firing a single musket.From that time forth the King was called upon to appear as an interested party in all the revolutionary scenes, and it was necessary to give him a part in every disturbance. Every hour had its dangers.It was necessary, too, that he should give some proofs of his firmness,even at the expense of his dignity. It was, therefore, necessary that he should not only give advice, but also that he should execute it. He could not do so, however, without being placed in some difficulty and embarrassment. If he were to send an officer to the Assembly with a written order, as he did, he could not avoid the risk of having him killed; and if in the Assembly itself he issued a proclamation, the magistrates could not fail to take notice of it, and would assuredly refuse him the opportunity of showing his strength. He therefore thought it necessary to put forward a bold step to enable the King to save his kingdom. He gave orders to go and see General Bugeaud, who commanded the French troops at his command.Bugeaud was very popular. His name was known to the nation, but not much known to the King. The King, on his part, had been very well known, and had been very favorably noticed at a time when the people of France were filled with anxiety for the safety of the crown. He went to him and spoke to him of this event, of the conduct of his forces, of the danger which threatened France, and of the imminent danger of his Majesty. Bugeaud saw that he was right, and did not hesitate; for there was no longer any need of saying, or of looking about, or of any sort of hesitation."I was at your Majesty's service," said he, "and I shall take care that you may not be obliged to regret it." He showed the King, by all the means in his power, that he considered the situation too dangerous to be abandoned, and that the only thing to be done was to carry the matter boldly through, without the slightest show of timidity. The King returned to Paris, and then Bugeaud marched for the scene of action.The town of the Faubourg St. Antoine still occupies a position surrounded by a double row of hedges, in which there are always sentinels placed to watch the approach of the inhabitants. It was through the gates of these hedges that the King and the deputies retired; but still it was necessary for them to pass through the streets to regain the town. They traversed these streets, the King being in advance of the others to take possession of the place. He was a magnificent specimen of a man,full of the vigor of youth and health, and with the strength of a Hercules. A great deal was said in the streets about his majesty,and they described a portrait of him in the character of Coriolanus.The King was accompanied by a numerous and splendid escort of the most distinguished persons--members of the Assembly and foreign ministers. The populace, eager once more to see a king whom they had so long adored, came out, from all directions, in bands, to meet them.They formed in two long lines along the streets; they crowded so closely behind the King that it was with the greatest difficulty that he was enabled to reach his dwelling. They came thither tumultuously, and they presented to him, not flowers, or wreaths, or any other tokens of adulation, but those tricolored cockades which are the emblem of the revolutionary power, and which the King was well aware how fond they were of. He could not refuse them, and, after having taken leave of them cordially, he left them rejoicing and contented.In the meantime the King proceeded to hold a session at the Tuileries.The Assembly had reassembled, and had made him a new proposition, if such it might be called, and the King had to determine what he was to do with it. He had already given his consent to the removal to Vincennes of those deputies who were still in Paris

\subsection{Sample - III}
the first time the subject was presented to me was at the house of a friend of mine named W. H. Green, whose father, at a dinner of his relations, the Barings, asked him if he ever read anything. The book he chose was Bulwer's romance, _Pelham_. The latter he read, and was highly gratified with its merits. Having become the possessor of this treasure, he determined to attempt a similar attempt on his own account. He therefore wrote out a dramatic _scena_, and went to the theatre to ask me for an introduction to Messrs. Sheridan and the Hon. Mr.Norton, whose company he then represented in the _Stranger_--a piece which came out at Drury Lane in the summer of 1822. The introduction,however, was not so readily obtained as he expected; the manager objected to the character of "Emilius," and the actor who supported him said that it would have been a great advantage to have given him his choice. On these representations Mr. Green made up his mind to write a play on the principles of Bulwer's _Pelham_; and, after an interval of about three months, produced his play, _The Adventures of Major de la Motte_. The acting of these two dramas was about all he had to bestow; the public, however, was abundantly satisfied with one of them,for it brought into general notice a very clever young man, at the then head of our profession, Edmund Kean; and the public were by no means displeased with the style of the acting of the other in which his brother-in-law, Mr. Green, was conspicuous.These plays had been represented to Mr. Green, at whose suggestion the tragedy written for him had been rejected, when I met him unexpectedly at the house of a friend, a few days after the conclusion of these performances. I was surprised at the warmth he manifested when I told him whom I had seen, of my own failure in the _Stranger_ case, and in his subsequent successes. He was delighted with the latter, but told me he feared the former had not been altogether satisfactory from a literary point of view.I was delighted however, when I read the play with him, he said, and immediately became enthusiastic in praise of the performance. He urged me the more to undertake more of such parts as Mr. Kean had so well filled, and even offered to give me two or three hundred pounds for the parts, in addition to any little salary I might think I should derive from the performance. I did not wait for his proposals to go further, but at once commenced writing out, preparatory to acting, the parts he had himself assigned to me. This step was not one that at first met with any opposition on the part of the actors of the company, but afterwards, as they found reasons to dislike the idea of my acting in any but their favourite characters, the affair took so serious a turn that the manager felt called on to interfere to prevent its being carried into effect. After some altercation with him,the matter was brought to a compromise, by the agreement that I,instead of retaining the character, was to give up the play to the company, at their own option, and that Mr. Kean was to assume the part of Sir Giles Overreach.When this piece was finished, and given to be acted at Drury Lane Theatre, by the company then in London, I was very nearly leaving it without seeing it, but I felt the importance of a rehearsal, so that the actors might be more ready to read it afterwards. Mr. Kean,however, who for some time had taken the play by way of a pattern,determined to proceed with it to the other theatres, and with a view to making it perfectly familiar, made me sit down with him to receive and read over the parts, that he might put down in my notes what alterations he thought advisable. It was arranged that he and Mr. Green should make their first appearance, with Mr. Kean to second him in Sir Giles Overreach. During the progress of the rehearsal, Mr.Kean requested Mr. Green to sit down on a chair I had borrowed to write down the character with, and to read it over in a distinct voice. It was a trying moment for two men like them, to start so diametrically opposite to each other in their parts. In the part of Sir Giles, Mr.Green was very nearly equal to Mr. Kean, having a good deal of natural power. It was as a _listener_ that Mr. Green won Mr. Kean's heart. When,therefore, Sir Giles made one of the speeches which had so excited my admiration at Drury Lane, Mr. Green listened with all the interest of a_listener_, but at the same time with a certain sarcastic curl of his lip. When he came to another, however, he was altogether the _listener_of the play, and his part was the _listener_ in this instance with a spice of the _speaker_.It was a difficult task to Mr. Kean to play a part with so much character in it; and in his hands I have seen Mr. Green put on a _whole host_ of characters in a minute. It used to be said of Mr. Kean's acting,that it was a _whole library_ of characters, and to hear him read a part over, was, for me, to begin with learning the scene to read it with him, and then the whole of it in its several parts. In the days of my youth, his reading was, at times, as interesting to me as any story-telling I ever listened to, and I never heard his readings through without feeling highly satisfied with myself for being an attentive listener to him. Mr. Kean never read a part over with me; indeed, as far as my memory serves me, he did not utter to me a single part of it aloud. After the first night it was not necessary that we should agree on the parts of Sir Giles. There the _listener_ (whose part, in this one instance, was not a difficult one to him) was more than a match for Mr. Kean; but from this time, and for several nights afterwards, the latter was in the habit of reading the part over in his usual manner, I being generally present. During this period, I was not so attentive as I otherwise should have been to Mr. Kean's readings; but I was so fascinated with them, that I never for an instant doubted that they afforded me the most intense enjoyment. If I was particularly fond of any scene, I used on more than one occasion to read it half aloud to the play-acting manager; and, as I could never overcome what was then in my voice a defect of hearing, I was frequently rewarded by hearing the tones of Mr. Kean's voice, with the accents I have just mentioned, coming from the other end of the theatre, when no person seemed to know any thing of its origin.Mr. Kean had a much longer and more difficult task than his brother in getting a play played, for Mr. Kean, after a certain stage success,was forced to give up everything as hopeless. In the autumn of 1847, he was engaged again to play for Messrs. Oxberry in the "Widow Married," which he did on the 16th of January, 1848; but that season, with the exception of one evening, was one of great fatigue to him. He gave up the stage for this engagement, as he said, to "have his hair cut," and this I believe he did, his grey locks being then closely clipped. In Mr. Kean's account of the following circumstances, he speaks of"this hair cutting" being a scene to which he refers on one occasion,saying, "If it had been my hair I should have got more satisfaction from my barber's art than from my razor;" and he mentions the following remark made in allusion to the incident:--"'How's this?' says Mr.Kean, as soon as the operation was over; 'this is a great loss.' 'Oh!yes, sir,' says the fellow; 'I know how little money I get for cutting a gentleman's hair; but I can cut your wig with ease; but your hair's a credit to the shop.'" Mr. Kean himself seems to have been aware that he was no longer so efficient in managing the part of a hero, as in his youth, and that there were times when he was really unable even to represent the characters suited to his talents. So it came to pass that the part of Sir Giles was handed over to Mr. Kean's brother, who gave up the other four. It may be imagined that the task of acting Sir Giles had not in this case been very light.While acting the part of that character he had to play the part of_Edmund_ to Mr. Kean's father, who had given me permission to give his story as I find it in Mr. Kean's manuscript:--"I had the honour of acting on one occasion at Drury-lane with Mr. Kean, who had the honour to be a pupil of Mr. Kean's, at Colebrook Street, Covent Garden,and the theatre had been closed in consequence of the non-performance of my _debutante_. I had the honour of appearing in my professional character; my name was made known to the audience; the manager sent for me and told me to go to the box which had been reserved for me the night before. I saw the box door open, and I entered it in triumph, and I found the occupants of it the great Mr. Kean and Miss O'Neil. No words can convey to your readers any idea of the triumph that was given to me. They introduced me to Mr. Kean, and the manager sent me to the theatre in the evening, and the curtain was drawn up on the last act of 'The Hunchback,' when Mr. Kean and Miss O'Neil made their _debut_on the stage. They were not long in creating a sensation. There were murmurs of applause that could be but one opinion as to their powers.The moment Mr. Kean had finished, there were cries of 'Mr. Smith! Mr.Smith!' and it was quite evident that he had been acting in his own name,and not in that of Mr. Kean. The actor's name was pronounced in a loud, decided tone, not the faint, piping cry of his brother-in-law.The effect was extraordinary; from this moment I was sure of Mr. Kean and his sister, and ever since has been my pride and my reward. Of course, if I had to be a manager myself, I should make it my business to look immediately into the merits of each one of these performers. I say that to my mind, the two were not equal for the purpose of the piece.Edmund Kean was the more powerful. There was a nervous motion, and a manner altogether superior to Mr. Kean, a great deal more majestic and impressive. He spoke more and better. Mr. Kean spoke in a louder, and, in my opinion at least, a better tone than the other; it was less that of an effeminate, than that of a manly actor."In the following letter to my father, I find Mr. Kean speaking of himself, in the _roles_ of Sir Giles Overreach in the _Courier_ and Sir Giles Overreach in the _Winter's Tale_, as follows: "At the close of the first act of the _Winter's Tale_, I entered into conversation with an acquaintance of mine. When I first saw him, in one of the boxes, it was evident that I was going to do him an injustice. I asked him to come down with me to the stage-door. He was absent at the moment, being occupied with an elderly lady, who was on her way to her carriage. I was not, however, so much astonished at his non-attendance, as was his mother; and I had learned, in the course of my professional acquaintance, that this venerable lady did not often alight from her carriage to walk about behind the scenes with her son. With her, he had been in the habit of making short, hurried visits, and with her, I could easily discern, that the mother had been in the habit of making short visits, and with her, the daughter had been in the habit of making short visits, and that both equally were in the habit of having short visits made to them."Such was Mr. Kean's manner, when he was at Exeter, in the year 1817; so changed by his residence in Paris, that the man who was the most accustomed actor of the two, now appeared the least so. Before I speak further of his first acting in London, I will give a sketch of his character on the stage, as it was at the opening of the theatre in 1809,at the Lyceum, in that city, on the 25th of January.A great actor, I have heard, in his more matured hours, can take pleasure in criticising the young efforts of his actors; and if any one doubts my statement, let him try the experiment. I myself do not think such an occupation necessary; but when it _is_ required, when no actor can perform his parts adequately, I should not be a little astonished if, in the character of Mr. Kean, he should not say with the poet:--"What, I think, I do,My actor can't tell;Perhaps I shall be An able man after all."But Mr. Kean's character on the stage at that time, consisted more in his acting than in anything else. He was the first manager who tried to put the best in the best place. He called his actors together,and said, "Now there must be no mistake about you, my hearties!" and then he would begin his remarks in this fashion: "This play is not for you, but for Mrs. Siddons; it is meant to show how the young men of this country must act. Do not let us, poor actors, be afraid of being laughed at and made to speak to a stupid, noisy town audience. They_are_ stupid, certainly; but they always laugh at you, and make a fool of you." It was this kind of thing that made Mr. Kean so admired,even in the midst of his success at Covent Garden; but the impression made upon us by his acting during Mr. Kemble's performances,when compared with that of Mr. Kean, is very different. At first, I thought him more agreeable; then I thought him more impressive, as he became better acquainted with the ways of the stage.We have here, on his arrival from Paris, the following letter from Mr. Kean:--(Received from Mr. Kean, on the 11th of December, 1812.)"MY DEAR SIR,"The theatre does not open until to-morrow evening, as I am anxious that it should be ready for the public when I return. It is the last public play in

\subsection{Sample - IV}
White-deer a pair of grey, northern Algonquin, also white-deer of a paler colour than common. Great babbler, the commonest of summer warblers,all these are found in a great number of localities in southern Ontario;but at Lake Erie and Lake Ontario, where they are few, they are quite common.Then, again, during the migration season they will often be seen consorting with their relatives the Canada Jay. On this account, a very large number of hawks that, though they are not regular songsters, are generally taken on the wing. But they are especially abundant in Newfoundland in the neighbourhood of the Little Fête and other great feasts, and are likewise met with in Newfoundland in winter, where they may be seen all the time, though they do not come in great numbers into the towns.Audubon tells us that although nearly all these birds spend the summer in Canada, yet they frequently winter in South America. Such have been frequently seen, but never described, by other observers. In studying any of these little northern warblers, we must go back to the winter quarters of these little birds, or at least see where they pass the summer.[Illustration: AMERICAN GOLDEN PLOVER, MALE AND FEMALE]How beautifully speckled are the breasts of these Golden Plovers! how beautifully spotted the upper parts of the head and breast, especially the under wing coverts. But on this account, their bright colours are particularly attractive, because the group is very abundant, and their close relative, the Golden Plover, is also frequently seen in the far north.This bird breeds sparingly in various parts of North America, but almost exclusively in Labrador. There it nests in small colonies of a dozen or more, making choice, I have no doubt, of some open, dry piece of ground,building their nests of grass and scraps of grass, placing them in the midst of grass on which, in company with their kindred, they pass the winter. The nest is built, in all probability, on the ground, or on the top of a tussock of grass or a tuft of oats, which has been dried, or rolled into a conical shape by birds, but which they have neglected to do for themselves; and after laying their eggs, they scrape down the soil upon which the nest is built, and together, with a few young, feed them all the summer. They pair about the end of April, and begin to breed so soon as the breeding season has passed, at the same time that the male bird may be seen sitting upon the outside of the nest.The nest of this species is not built as closely as that of the English species, and not being peculiar to America, a large number of its eggs has been obtained in Great Britain, and it is highly probable that it exists abundantly in the United States also.In breeding time, the Gulls and Terns, as well as the other birds, do not congregate in large flocks, but generally avoid flocks that are daily passing, and thereby contribute very much towards diminishing the number of their feathered associates, which, being fewer, would be more easily preserved. The same thing may be said of the very numerous young which come with the large migration northward, and, in a measure,counteract the tendency to overcrowding.But although the Gulls and Terns are thus apt to resort to the north in winter,how many of the same species are known to breed in the other parts of the world? The British Islands, indeed, are but thinly populated, and the season for breeding does not arrive so early as that for breeding in Europe. We find, therefore, in the British Islands only a few pairs or very few individuals. The Skuas and Petrels are probably more numerous,but such is the local distribution of this species, that it is difficult to find more than three or four of its breeding haunts. Our only figure of this species is in the "Manual" for the year 1858, in which it is figured under the name of _Crex pusilla_.[Illustration: BLACK GUILLEMOT]BLACK GUILLEMOT.*       *       *       *       *SPECIFIC CHARACTER.BLACK GUILLEMOT.--Bill, the base of the upper mandible and the tip of the ear black; legs, legs, toes, and feet, black; wings, blackish, the feathers margined with dull ash-grey; upper parts ash-grey; quills blackish, margined with greyish; tail blackish, the inner three feathers of the outer web tinged with brown, and the next tipped with white, except on the inner web; the two outer feathers of the outer web tipped with white.*       *       *       *       *The present species was discovered by Captain King at Sitka, in Russian America, and may be distinguished from the preceding by its black rump,beneath which are eight blackish-brown lines, beginning at the base of the feathers. In its haunts, it is rather tame, but in autumn it seldom perches on trees. On the coast the breed begins to breed in December, and by the end of April it will have laid about six eggs. It is somewhat gregarious, sometimes in large flocks. A female caught in Baffin's Bay in 1825 was of a sooty black colour above and light ash-grey below, with three of the tail-feathers of a blackish tinge.*       *       *       *       *TEMMINCK'S GUILLEMOT.TEMMINCK'S GUILLEMOT (_Haematopus bairdii_) is said to have been taken near the mouth of the Columbia, and by Captain Cook has been called the Common Guillemot.TEMMINCK'S HELMET.TEMMINCK'S HELMET. Plate XXI. fig. 3.*       *       *       *       *Adult Male. Plate XXII. fig. 1, 2.Bill, the base of the upper mandible and the tip of the ear black; legs,feet, toes, and feet black; upper part of the head and neck dark ash-grey;back, scapulars, wing-coverts, and quills black, the latter margined with pale greyish-white; tail of the same colour, the middle feathers of the outer web at the end tipped with white; three outer feathers of the same, and the next two very slightly tipped with the same; lower parts white.Total length 5 inches, extent of wings 5, depth of body 2 1/2 inches.This species is only two feet ten inches in length, and during the summer time, during which it can be seen floating on the ocean in autumn,resembles the preceding, but it is so extremely scarce, that it is rather a difficult matter to ascertain its haunts. I have no doubt that it migrates from Europe, across the Atlantic, to the north, even where it is now known to be extinct.*       *       *       *       *AMERICAN SEA-EAGLE.EIDER-BILLED BOOBY.*       *       *       *       *_HaliaA|etus leucogaster_, Wils.*       *       *       *       *AMERICAN WHITE-FRONTED BOOBY (_HaliaA|etus leucogaster_, TEMM.) is one of the smallest of the American species, measuring only five inches and three quarters in length. The bill is black, and the feet deep brown. It is a bird in the collection of the late Mr. John Cassin of New York, and was shot in the neighborhood of Lake Erie. Length 5 inches and 3/4, extent of wings 3 inches and 1/4, depth of body 1 1/2.*       *       *       *       *I have been indebted for the above description of the Blue-headed Buzzards to my friend, Mr. Wm. L. Beal.*       *       *       *       *PALL MALL BLUE-HEADED BOOBY (_HaliA|etus pallens_, TEMM.) may be distinguished by the reddish band over the eye, and the brown patch on the primaries,which are longer and more attenuated, than the black ones of the last species, the bill being a little broader and red, and the legs lighter than those of the last species. It has been called the Alpine Blue-headed Booby, by the late Dr. Edward Smith, in his description of this bird. I believe that there is but little difference in its appearance, except the colour of the bill, which in the male is of a dark brown, in the female yellow.*       *       *       *       *HORNED OWL._Strix flammea_, LINN.*       *       *       *       *_Strix argemone_, LINN.*       *       *       *       *The habits of the Horned Owl are, like those of the Snow Owl and the Long-eared Owl, imperfectly known. They have long been familiar objects to the inhabitants of the northern parts of our country, who are accustomed to their appearance and mode of travelling in companies. They are most frequently seen in the night. It is often heard to hoot, or squeal,and at times is very noisy.It is found during the whole of the northern summer, on the pine plains and barrens, on the <DW72>s of the higher elevations of our country, and in the northern parts of Maine, Nova Scotia, Newfoundland, and in several parts of New England. It is one of the most common inhabitants of our villages, and is so extremely restless and active, that it is almost impossible to catch it. They are very bold and noisy, rising from the tops of the low bushes and branches, and making a terrible hissing, as they do when alarmed, which will draw on them the attention of the person who perceives them. They are generally seen in flocks, and at all times wary, giving notice of the approach of danger, by their peculiar crowing, and various notes, which are peculiar to themselves, and often mistaken for a call. Their note resembles that of the Owl, and is much louder, resembling the cry of the Great Horned Owl.*       *       *       *       *I have been thus particular in giving you the above description, as I believe this species to be the one I have already figured. You will readily believe that it would be impossible for me to decide in which of the two localities which I have described the bird is to be looked for. I only mention the latter, as the description agrees better with that of the present bird than with that of any other in which I have seen it.*       *       *       *       *CHIMÆOLURUS VIRGINIANUS, _Lath._ Ind. Ornith. vol. ii. p. 301.—_Ch.Bonaparte_, Synops. of Birds of the United States, p. 54.CHIMÆOLURUS VIRGINIANUS, _Nuttall_, Manual, part i. p. 215.AMERICAN CHIMÆOLURUS, CHIMÆOLURUS AMERICANUS, _Ch. Bonaparte_, Amer.Ornith. vol. ii. p. 39. pl. ii. fig. 2.—_Nuttall_, Manual, p. 209.Adult Male. Plate XXIII. Fig. 1.Bill rather long, slender, strong, compressed toward the end; upper mandible with the dorsal outline a little convex, the ridge rather wide and flat,the sides convex from the base, the edges overlapping, the tip declinate;lower mandible with the angle narrow and very long, the dorsal line rather convex, the sides rounded, the tip acute. Nostrils basal, lateral,round, covered by the reversed filaments of the frontal sinuses. Head rather large. Body moderate. Legs of ordinary length; tarsus very strong,scutellate anteriorly, acute behind; toes free, scutellate above, the lateral ones nearly equal, the hind toe larger; claws of ordinary length,compressed.Plumage soft, blended, somewhat blended, not glossy. Wings rather long,third quill longest, second and fourth equal. Tail of ordinary length,slightly emarginate, the two lateral feathers longest, the two lateral inferior with some small tips.Bill deep brown, black at the end, paler at the sides. Iris brown. Feet flesh-colour. Head and neck pale ash-grey. Back, scapulars, and rump dark umber-brown, reflecting into deep brown, the tail, secondary quills,and coverts, as well as the ends of the secondary quills, and tips of the larger ones, white. Wings dusky, their coverts margined externally with reddish-brown. Fore part of the back, breast, and abdomen deep brown, tinged with orange; the breast tinged with yellow, the abdomen with a tinge of dull red. On the breast a broad band of dusky red on each side.Length 7 inches, extent of wings 10; bill along the ridge 1-3/12, along the gap 1-1/12; tarsus 2-1/12.Adult Female. Plate XXIII. Fig. 2.The Female resembles the male, but somewhat resembles the white-headed Woodpecker, the head, neck, breast, and abdomen being pale ash-grey.The young resemble the female, and differ from the male, in having the chin and fore part of the breast light ash-grey, and the rest of the under parts ash-grey.THE COTTON PLANT.GOSSIUM GLYCYLLARUM, _Willd._ Sp. Pl. vol. ii. p. 779. _Pursh_,Flor. Amer. vol. ii. p. 422.—DECANDRIA MONOGYNIA, _Linn._DECANDRIA RHAMNACEAE, _Juss._This plant, from which the generic name of this genus is derived,is distinguished by its pendulous cymes of large, silky, terminal panicles, and by the sinuosities of the branches, which are mostly smooth. The leaves are cordate, downy, and attenuated at the base. The flowers are pale orange-, and exhale a strong and very pleasant odour.THE HIGH BERRIES OF THE NORTH.(_MAGNOLIA CANADENSIS_, DESK.) NORTH OF KINGSBRIDGE.[Illustration: THE HIGH BERRIES OF THE NORTH.]The highest trees in the county of Brunswick are found near the town of Kingston; but the low and more sheltered parts of the country have abundance of the low-growing aromatic, which grows there from seed, and is,consequently, of a superior quality. Not more than fifty or sixty miles below the town of St. John's, this shrub attains a height of upwards of fifty feet, with spreading branches of beautiful spreading foliage.THE CRANE CRANE._CATHARTE CANADENSIS_, TEMM.PLATE XXIII. MALE AND FEMALE.This species has never, or very rarely, been observed on our seaboard during the spring and summer, unless I mistake not, as is said by the natives, in many parts of Newfoundland. It frequently comes within a few miles of the sea-shore, and after passing over the downs or beach, settles upon the marshes or small islands, erecting its nest on the summit of a large tree, and generally resting on the trunk. There is, at all times, a sufficient number of young ones to fill its nest, and, consequently,it seldom requires to be robbed. It generally dwells upon high and exposed situations, yet never in an open forest. As many as four or five nests of this species may often be observed on a single tree, situated on a level with the ground, or where the lower branches have been broken off by storms. It sits upright, with its neck or tail drawn in, and so rarely, on opening its mouth, that you may often look down into it, and take your bird out by the neck or tail. It is only during the autumn, and towards the close of that season, that it deserts the salt marshes, retires to its aerial breeding-places, and generally makes its nest on a swamp or river island. The habits of this bird are so like those of the common stone crane, that it would have escaped notice were it not for the variation in the colour of its bill. This is of a white colour, shading off towards the tips of the upper mandible, which are pale brown.So common is this species on the Atlantic seaboard, that few persons can fail to have seen it. While on board our ship at St. John's, on the 30th of October 1828, I noticed many of these birds on a small pond that runs near our town. They were wading about and darting from one point of the shore to another, as if searching for a distant fish. They were rather shyer than the common white crane, but had the same abrupt note,so different from that of the red-necked species. They continued to hop about the pond, looking out for food, the whole time that the vessel remained there.
\end{document}